\documentclass[10pt,twocolumn,letterpaper]{article}

\usepackage[pagenumbers]{cvpr} % To force page numbers, e.g. for an arXiv version

\usepackage{url}
\usepackage[utf8]{inputenc} % allow utf-8 input
\usepackage{booktabs}       % professional-quality tables
\usepackage{amsfonts}       % blackboard math symbols
\usepackage{nicefrac}       % compact symbols for 1/2, etc.
\usepackage{microtype}      % microtypography
\usepackage{xcolor}         % colors
\usepackage{multirow}
\usepackage{dsfont}
\usepackage{soul}
\usepackage{adjustbox}
\usepackage{xspace}
\usepackage{bm}
\usepackage{bbm}
\usepackage{amsmath}
\usepackage{anyfontsize}
\usepackage{algpseudocode}                  % algorithm formatting
\usepackage{setspace}                       % pleasant display of algorithm.
\usepackage{scrextend}
\usepackage{mathtools}
\usepackage{graphicx}
\usepackage{subcaption}
\usepackage{enumitem}
\usepackage{colortbl}
\usepackage{wrapfig}
\usepackage{kotex}
\usepackage{lipsum}
\usepackage{xcolor,pifont}

\usepackage{duckuments}
\usepackage{makecell}
\usepackage[ruled,vlined]{algorithm2e}
\usepackage{flushend}
\usepackage[usestackEOL]{stackengine}
\usepackage{listings}
\usepackage{minted}
\usepackage{inconsolata}
\usepackage{xcolor}
\usepackage{arydshln}
\usepackage{booktabs}
\usepackage{multirow}
\usepackage{pgfplots}

\usepackage{tikzducks}
\usepackage{graphicx}
\usepackage{wrapfig}
\usepackage{lipsum}

\definecolor{cvprblue}{rgb}{0.21,0.49,0.74}
\usepackage[pagebackref,breaklinks,colorlinks,allcolors=cvprblue]{hyperref}

\def\paperID{10159} % *** Enter the Paper ID here
\def\confName{CVPR}
\def\confYear{2026}

\title{Rethinking Prompt Design for Inference-time Scaling in Text-to-Visual Generation}

\author{Subin Kim$^{1}$ 
\qquad Sangwoo Mo$^{2}$ 
\qquad Mamshad Nayeem Rizve$^{3}$ \\
\qquad Yiran Xu$^{3}$ 
\qquad Difan Liu$^{3}$
\qquad Jinwoo Shin$^{1}$
\qquad Tobias Hinz$^{4}$\\
$^{1}$KAIST 
\qquad $^{2}$POSTECH
\qquad $^{3}$Adobe
\qquad $^{4}$Meta
\\
{\tt\small subin-kim@kaist.ac.kr}
}

\newcommand{\flname}{Prompt Redesign for Inference-time Scaling\xspace} % long name 
\newcommand{\fsname}{PRIS\xspace} % short name

\newcommand{\vlname}{Element-level Factual Correction\xspace} % long name 
\newcommand{\vsname}{EFC\xspace} % short name

\begin{document}
\maketitle
\begin{abstract}
Achieving precise alignment between user intent and generated visuals remains a central challenge in text-to-visual generation, as a single attempt often fails to produce the desired output. 
To handle this, prior approaches mainly scale the visual generation process (e.g., increasing sampling steps or seeds), but this quickly leads to a quality plateau. 
This limitation arises because the prompt, crucial for guiding generation, is kept fixed. 
To address this, we propose Prompt Redesign for Inference-time Scaling, coined PRIS, a framework that adaptively revises the prompt during inference in response to the scaled visual generations. 
The core idea of PRIS is to review the generated visuals, identify recurring failure patterns across visuals, and redesign the prompt accordingly before regenerating the visuals with the revised prompt.
To provide precise alignment feedback for prompt revision, we introduce a new verifier, element-level factual correction, which evaluates the alignment between prompt attributes and generated visuals at a fine-grained level, achieving more accurate and interpretable assessments than holistic measures.
Extensive experiments on both text-to-image and text-to-video benchmarks demonstrate the effectiveness of our approach, including a 15\% gain on VBench 2.0. These results highlight that jointly scaling prompts and visuals is key to fully leveraging scaling laws at inference-time. Visualizations are available at the website: \url{https://subin-kim-cv.github.io/PRIS}.
\end{abstract}
\section{Introduction}
Generative models~\citep{comanici2025gemini, flux2024, wan2025} have achieved remarkable progress across various domains, including language, image, and video, demonstrating strong capabilities in modeling complex data distributions.
In the visual domain, denoising models~\citep{ho2020denoising, lipman2023flow} conditioned on textual prompts now allow users to generate high-quality images and videos directly from natural language.
However, as prompts become more intricate, e.g., requiring compositional structures in images or complex motion, camera movements, and causal orders in videos, it becomes increasingly challenging to obtain outputs that fully align with the prompt in a single attempt.

Recent work addresses this shortfall in text-visual alignment by allocating additional compute at inference time (i.e., inference-time scaling).
These approaches typically scale the visual generation either by increasing the compute budget for decoding a single candidate from a prompt~\citep{ma2025scaling}, or by generating multiple candidates for the same prompt to produce a diverse pool of visual outputs~\citep{kim2025inference, kim2025testtime, he2025scaling}.
However, they primarily focus on scaling visual parts while keeping the input prompt fixed.
This creates a key bottleneck because many generation errors arise from ambiguous or incomplete prompts, and scaling visuals conditioned on a suboptimal prompt offers limited benefit since the prompt provides essential guidance for conditional generation.

\begin{figure*}[t!]
\vspace{-0.1em}
\centering
\includegraphics[width=0.99\textwidth]{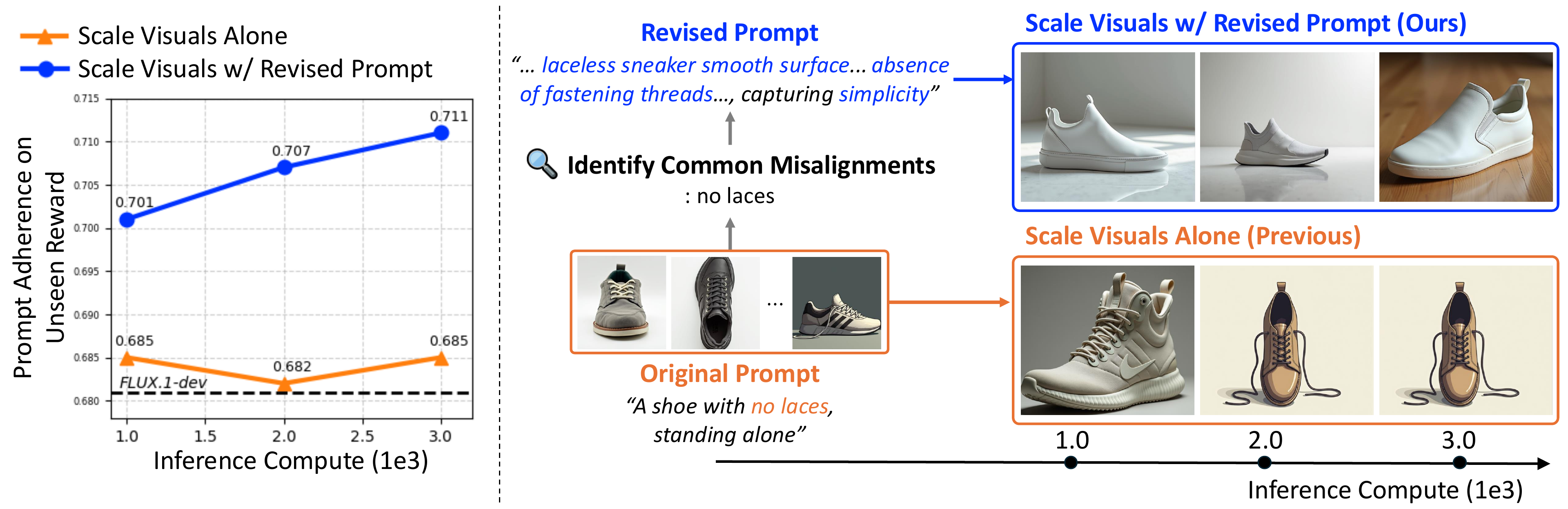}
\vspace{-0.12in}
\caption{\textbf{Our prompt redesign scales with compute, while fixed-prompts plateau}.
Given a user-provided complex text prompt, scaling visuals alone with a fixed prompt at inference time often leads to early performance plateaus, especially for unseen rewards (see \textcolor{orange}{orange} line and boxes). It also repeatedly produces outputs that exhibit common failures and cover only parts of the prompt, even as compute increases to sample more visuals. In contrast, scaling visuals alongside our redesigned prompts yields progressively improved generations and substantially higher prompt-adherence scores as compute increases for both given and unseen rewards (see \textcolor{blue}{blue} line and boxes).
}
\label{fig:main_figure}
\end{figure*}

More importantly, visual scaling consistently reveals recurring generative failures.
For example, in Figure~\ref{fig:main_figure}, when scaling with the intent ``a shoe with no laces, standing alone,'' the element ``a shoe'' is consistently achieved, yet laces still appear in every output.
These failure patterns become even more pronounced as prompts grow more complex, such as in text-to-video generation, where producing a high-fidelity sample becomes substantially harder.
However, prior prompt-refinement approaches~\citep{brade2023promptify, wang2024promptcharm, datta-etal-2024-prompt, hao2023optimizing} are confined to individual samples, focusing on sample-specific deviations.
As a result, they fail to correct the recurring failure modes that consistently appear across samples when scaling, missing the opportunity to jointly improve both the conditioning text and the generated outputs.

% by revising the text prompt according to the generative patterns revealed through scaled visuals

To address these limitations, we extend inference-time scaling beyond the visual domain to the text prompts, proposing Prompt Redesign for Inference-time Scaling (\fsname). 
Instead of passively waiting for a high-scoring sample when scaling visuals, \fsname identifies recurring failure modes across scaled visuals and adaptively revises the prompt to reinforce commonly under-addressed aspects while preserving the user’s original intent.
Consequently, whereas fixed-prompt inference-time scaling quickly plateaus in prompt adherence even as compute increases due to repeated failures (see \textcolor{orange}{orange} line in Figure~\ref{fig:main_figure}), \fsname leverages compute more effectively by jointly scaling prompts with the scaled the visuals, enabling sustained improvements in text-visual alignment under the scaling law (see \textcolor{blue}{blue} line in Figure~\ref{fig:main_figure}).

\begin{figure*}[t!]
    \centering
\includegraphics[width=0.95\textwidth]{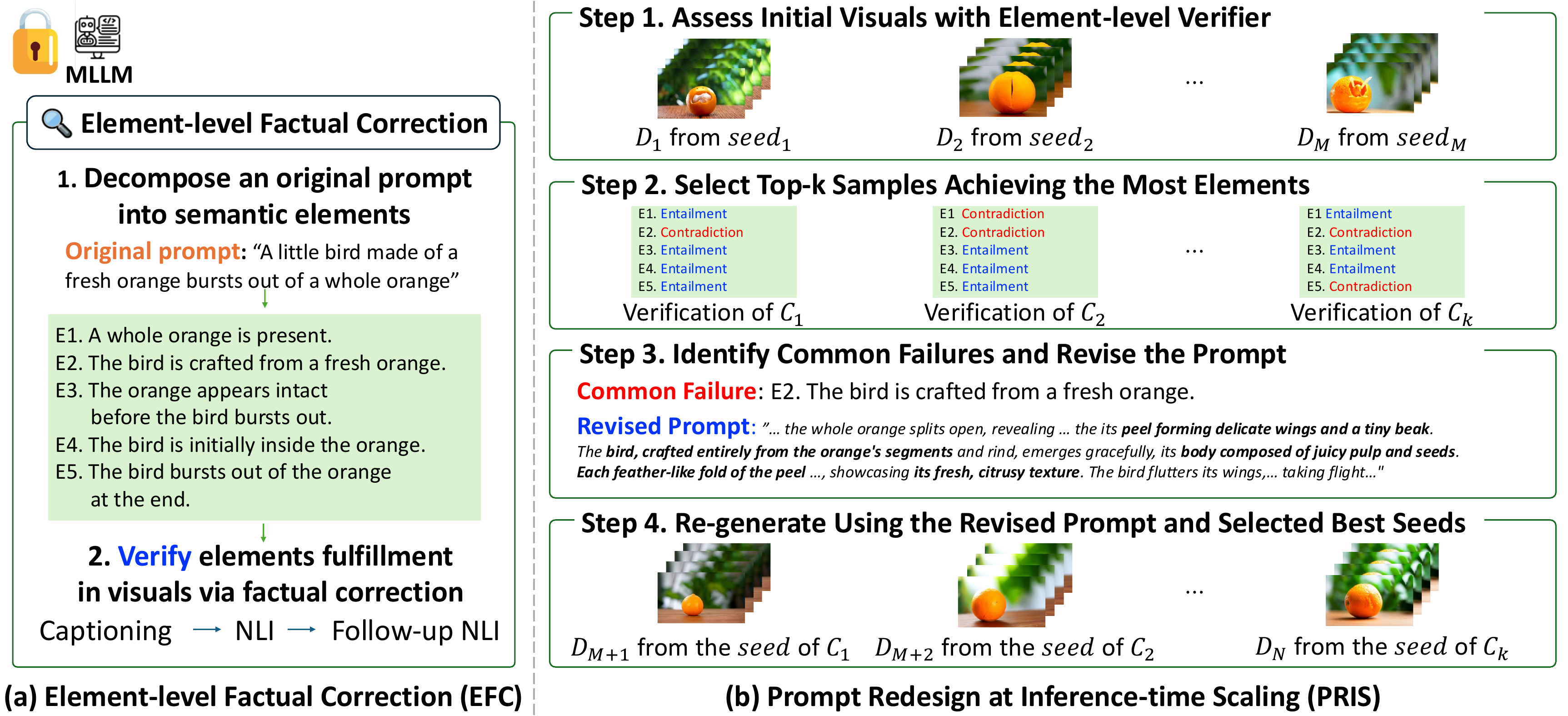}
\vspace{-0.1in}
    \caption{\textbf{Overview of \flname (\fsname)}, which leverages diagnostic feedback from our verifier \vsname to revise prompts during inference based on generated visuals. \vsname decomposes prompts into semantic elements and verifies each element for fine-grained text-visual alignment (left). Guided by the EFC, PRIS proceeds as follows (right): Step 1 reviews initial generations with \vsname; Step 2 selects top-$k$ successful samples and identifies recurring failures; Step 3 redesigns the prompt to emphasize common failures; and Step 4 regenerates visuals with the revised prompt and top-$k$ seeds. The process can be iterated by returning from Step 4 to Step 2.
    }
    \label{fig:framework}
\vspace{-0.2in}
\end{figure*}

To identify failure patterns for prompt revision, we develop \vlname (\vsname), a fined-grained descriptive verifier to examining the generated visuals, built on a multimodal large language model (MLLM) (see Figure~\ref{fig:framework}).
When assessing the alignment between the visuals and the prompt, \vsname first decomposes the prompt into disjoint semantic elements and verifies each against a caption of the generated visual, framing every element as a textual hypothesis.
This text-to-text comparison mitigates the affirmative bias common in MLLM‑based text-visual question answering~\citep{fu2025hidden, bai2024hallucination, han2024instinctive}, thus improving verification accuracy.
We further introduce a benchmark pairing each prompt with multiple generated visuals, some fully aligned, others only partially so. 
On this dataset, \vsname consistently distinguishes ground-truth visuals from plausible but misaligned distractors, significantly outperforming existing verifiers.

% \vspace{0.2em}
\noindent\textbf{Contributions.} Our contributions are as follows:
\begin{itemize}
\item We propose \fsname, \flname, which identifies recurring failure patterns during visual scaling and revises prompts accordingly.
\item We introduce a new verifier that assesses fine-grained alignments with text-based comparisons for prompt redesign.
\item \fsname consistently enhances text-visual alignment without compromising visual fidelity, yielding a 7\% improvement on GenAI-Bench for text-to-image and a 15\% improvement on VBench2.0 for text-to-video generation.
\item We present the first benchmark for evaluating verifiers in inference-time scaling.
\end{itemize}

\section{Related Work}

\textbf{Scaling inference-time compute in visual generation.}
Despite recent progress driven by powerful denoising architectures~\citep{ho2020denoising, lipman2023flow}, producing faithful outputs in text-to-visual generation remains challenging, particularly for complex prompts. 
Since outputs are shaped jointly by the initial noise, the sampling trajectory, and the prompt, existing inference-time scaling methods allocate additional compute to exploring better noise seeds and trajectories, either by increasing sampling steps or by generating multiple candidates~\citep{ma2025scaling}, often with advanced search algorithms~\citep{kim2025inference, kim2025testtime}.
Reward models~\citep{liu2025improving, zhang2025videollama} then score these candidates and select the best one, either at the final output stage (Best-of-$N$; BoN)~\citep{ma2025scaling} or during sampling through Search-over-Paths~\citep{he2025scaling}, which propagates high-reward trajectories.
However, all prior methods share a key limitation: they expand only the visual search space while keeping the prompt fixed. In contrast, we treat the prompt as an equally critical and previously underexplored axis of inference-time scaling. Rather than discarding low-scoring generations, we analyze their recurrent failure patterns and redesign the prompt jointly with visual scaling, enabling subsequent generations to receive progressively stronger and more targeted guidance.

\noindent\textbf{Prompt design in text-to-visual generations.} Prompt design is a critical component of text-conditioned generation that serves not merely as a pre-processing step~\citep{zheng2025vbench2} but as a means to improve model comprehension, output quality, and adherence to the input description.
Since the prompt itself guides the generation, even different phrasings of the same user intent can produce markedly different outputs.
Yet, crafting effective prompts remains challenging, often requiring tedious trial-and-error.
To address this, recent approaches~\citep{brade2023promptify, wang2024promptcharm, datta-etal-2024-prompt, hao2023optimizing, nailei2024uffgtg} propose systems that interactively help users explore alternative phrasings or automatically rewrite prompts, reducing reliance on na\"ive iterations.
However, these approaches typically require human involvement~\citep{brade2023promptify, wang2024promptcharm}, and more importantly, they operate independently of inference-time scaling~\citep{datta-etal-2024-prompt, hao2023optimizing, nailei2024uffgtg}: their per-sample rewrites react to individual noisy failures rather than addressing the recurring failure patterns that are revealed through visual scaling, limiting their ability to improve adherence.
To fill this gap, we demonstrate that redesigning prompts based on \textit{shared failures} across samples---rather than isolated per-sample corrections---achieves substantially higher adherence as compute scales. 
Our method applies to both text-to-image (T2I) and text-to-video (T2V) generation, whereas prior prompt-refinement efforts largely focus on T2I generations.

\noindent\textbf{Chain-of-thought and reasoning.}
Incorporating chain-of-thought (CoT) reasoning into visual generation has emerged as a promising paradigm for improving image quality through iterative reflection and guidance~\citep{wang2025mint, jiang2025t2i, liao2025imagegen, guo2025can, zhuo2025reflection}. 
Recent works pursue this direction via unified models~\citep{tian2025unigen} that combine visual understanding and generation and jointly optimize large language models with multimodal objectives and generation-specific losses.
Our approach differs in two key aspects.
First, we use the off-the-shelf MLLM~\citep{Qwen2.5-VL}, without any additional training, to scale prompts for arbitrary text-conditioned generators.
Thus, our design supports both T2I and T2V settings and remains compatible with unified models for prompt refinement.
Second, existing CoT-based approaches typically apply reasoning at the per-sample level, refining each output independently~\citep{zhuo2025reflection} or deciding whether a particular sample should be retained or discarded~\citep{guo2025can}.
In contrast, our method aggregates information across the samples generated during scaling and updates the prompt based on cross-sample trends rather than isolated reflections.

\vspace{-0.5em}
\section{\flname}
To enable common-failure-aware visual feedback in inference-time scaling, we introduce a new verifier, \vsname, which provides fine-grained assessments of generated visuals (Section~\ref{sec:method_1}). Building on these assessments, we then present our prompt redesign strategy, \fsname, which extracts common failures and revises the prompt accordingly to improve fidelity as compute scales (Section~\ref{sec:method_2}).

\subsection{\vsname: \vlname}\label{sec:method_1}
Our goal is to identify recurring misalignments between the original prompt and the generated visuals, focusing on elements that the generator repeatedly fails to realize, including missing components, incorrect causal relations, and disordered temporal motions.
Achieving this requires a fine-grained visual verifier capable of determining, for each generation, which prompt elements are satisfied or missing, as previous single-scalar alignment scores from previous verifiers~\citep{lin2024evaluating, liu2025improving} cannot reveal such details.
To this end, we introduce \vlname (\vsname), a new verifier that provides interpretable, fine-grained text-visual assessments using an MLLM without additional training. See Figure~\ref{fig:framework} (a) for an overview; further illustrations of \vsname are provided in Appendix~\ref{appen:benchmark}.

\noindent\textbf{Break down the prompt into fine-grained elements.}
Holistic alignment scores often obscure which aspects of a generation are actually satisfied; as prompts become more complex, a visual can miss some of the required elements, and such omissions cannot be revealed by a single holistic score.
To address this, \vsname first decomposes the original prompt $p$ into a set of verifiable atomic semantic elements $p=\{p_1, p_2, ..., p_s\}$, where each element $p_i$ corresponds to a distinct element.
Here, atomic facts are extracted according to predefined semantic categories, such as image-level elements covering object presence, properties, and spatial arrangement, and motion-level elements covering object motion, camera movement, scene transitions, and temporal ordering.
Then, \vsname classifies each $p_i$ as either $\{\text{core, extra}\}$. 
The core elements are objective, factual, and essential to the intent of the prompt, while the extra elements are more subjective or stylistic, so they are often flexibly interpreted. 
This classification is then used to guide the prioritization of generated samples during final scoring.

\vspace{0.1em}
\noindent\textbf{Assess each element through factual correction.}
After decomposing the prompt into multiple elements, \vsname performs factual correction on each element by evaluating it against each generated visual $D$ to determine whether the element is accurately realized.
Here, \vsname assesses alignment using a text-text verification approach, rather than direct yes/no visual question answering (VQA), because it achieves higher accuracy by mitigating confirmation bias and consistently providing interpretable descriptions, whereas binary VQA often omit such information.
% (detailed ablations are reported in Section~\ref{sec:exp_verifier} and Appendix~\ref{appen:benchmark}).
To enable this, \vsname first generates a natural-language caption for $D$, then infers the relationship between each element $p_i$ and this caption.
This step is formulated as a natural language inference (NLI) task: if the caption semantically supports the element, the relationship is labeled as \textit{entailment}; if the caption contradicts the element, it is labeled as \textit{contradiction}; and if the caption does not provide sufficient information to confirm the element, the label is \textit{neutral}.
For any element $p_i$ initially classified as \textit{neutral}, such as when the caption omits or ambiguously describes it, \vsname generates an open-ended question $q_i$ whose expected answer corresponds to $p_i$, without binary framing.
Then, \vsname queries $q_i$ over the visual $D$, obtain a free-form response, evaluate it against $p_i$ in a second NLI step, and relabels the element as either \textit{entailment} or \textit{contradiction}.

\vspace{0.1em}
\noindent\textbf{Prioritize core elements in final scoring.} 
After factual correction, we obtain verification results $C$ for each visual. 
\vsname then assigns a score based on the number of elements labeled as \textit{entailment} in $C$. Core elements are prioritized because they are objective, factual, and less open to subjective interpretation, making them essential for faithfully capturing the prompt’s intent. When multiple candidates achieve the same core accuracy, \vsname breaks ties using extra-element accuracy.

% Right now the process is described as a two-stage process: generate, then refine. In theory this can be repeated several times, i.e., generate, refine, refine, refine, etc. Maybe update the notation to make this clear and you can mention that, in practice, we only do it for two stages

\subsection{Common-failure-aware Prompt Redesign \\for Inference-time Scaling}\label{sec:method_2}
We propose a prompt redesign framework, \fsname, which revises the prompt to address common failures across samples using the fine-grained text-visual alignment assessments produced by our verifier \vsname. 
By pinpointing where alignment breaks down in earlier generations, \fsname incorporates these diagnostic signals into subsequent prompt updates, guiding later samples toward higher fidelity to the original intent. See Figure~\ref{fig:framework} (b) for an overview.

\begin{itemize}[leftmargin=1.em]
    \item \textbf{Step 1. Generation and verification.} \fsname first generates $M$ candidate visual samples and evaluates the fulfillment of elements $\{p_1, p_2, \dots, p_s\}$ for each sample using our verifier \vsname (Section~\ref{sec:method_1}), obtaining verification results for each sample ($C_{1}$ through $C_{M}$). 

    \item \textbf{Step 2. Select the top-$k$ best-performing samples.} \fsname then selects the top-$k$ samples that collectively cover the largest number of elements, with ties further resolved using the scalar score from a reward model~\citep{lin2024evaluating, liu2025improving} trained on human-preference datasets.
    This ensures that the selected samples better reflect human-preferred ones.
    
    \item \textbf{Step 3. Identify misalignment patterns and revise the prompt.} Within the selected subset, \fsname identifies common failures, defined as elements whose success probability is below 50\% within the top-$k$ samples, by aggregating the element-level assessments from \vsname across these visuals.
    Based on these common failures, \fsname revises the \textcolor{orange}{original prompt} $p$ into a \textcolor{blue}{revised prompt} $p'$ that explicitly reinforces overlooked elements while preserving those already well addressed.
    This targeted refinement encourages subsequent generations to focus more effectively on underrepresented elements.
    If no common failures are observed (i.e., every element has a success probability above 50\%), \fsname instead treats the prompt itself as the refinement target to encourage exploration of prompt variations.

    \item \textbf{Step 4. Regenerate with the revised prompt and selected noise conditions.}
    Using the revised prompt $p'$, \fsname regenerates $(N-M)$ samples by reusing the noise latents of the top-$k$ samples. Reusing these seeds preserves earlier successes more reliably than random initialization, as certain noise conditions are known to yield better alignment for specific prompt types~\citep{zhou2025golden, ahn2024noise, qi2024not}. After regeneration, \fsname verifies and ranks the samples using \vsname.
\end{itemize}

\vspace{-0.2em}
\begin{table*}[t]

\centering

\begin{minipage}[t]{0.45\textwidth}
\centering
\vspace{-0.15in}
\captionof{table}{\textbf{Quantitative results of T2I on GenAI-Bench.} $^{*}$ denotes results with standard prompt expansion; BoN refers to ``Best-of-$N$'' selection using fixed prompts.
\textbf{Bold} shows the best.
}\label{tab:quan_flux}
\vspace{-0.15in}
\resizebox{\linewidth}{!}{
\begin{tabular}{l ccc}
\toprule
\textbf{Method} & \makecell{VQA-\\Score\\(Given)} & \makecell{DA-Score \\ w. BLIP2-VQA\\(Unseen)} & \makecell{Aesthetic \\ Quality \\ (Unseen)}  \\

\midrule
FLUX.1-dev~\citep{flux2024} & 0.718 & 0.681 & 5.764 \\
\cdashline{1-4}\addlinespace[3pt]
$+$BoN & 0.783 & 0.682 & 5.761 \\
\rowcolor{green!7}
$+$\textbf{\fsname} & \textbf{0.854} & \textbf{0.707} & \textbf{5.765} \\

\midrule
FLUX.1-dev$^{*}$~\citep{flux2024} & 0.769 & 0.695 & 5.824 \\
\cdashline{1-4}\addlinespace[3pt]
$+$BoN$^{*}$ & 0.829 & 0.710 & 5.820\\
% $+$Rand. Prompt Refine & 0.829 & -- & -- \\
\rowcolor{green!7}
$+$\textbf{\fsname}$^{*}$ & \textbf{0.853} & \textbf{0.713}& \textbf{5.841} \\
\bottomrule

\end{tabular}
}
\end{minipage}
\hfill
\begin{minipage}[t]{0.54\textwidth}
\centering
\vspace{-0.2in}
\includegraphics[width=\linewidth]{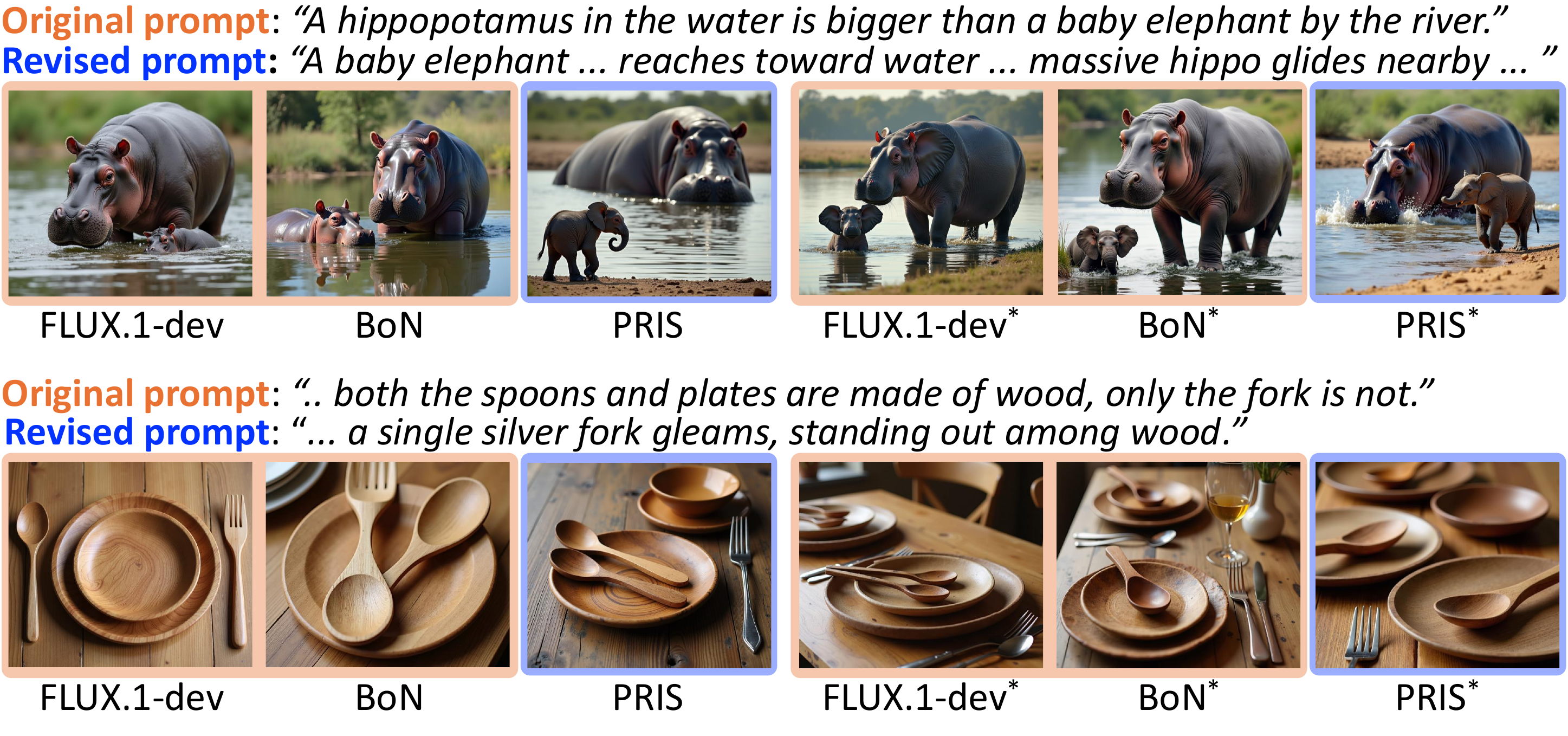}
\vspace{-0.3in}
\captionof{figure}{\textbf{Qualitative comparisons of T2I generation}. $^{*}$ denotes results with standard prompt expansion.
}\label{fig:qual_flux}
\end{minipage}

\vspace{-0.15in}
\end{table*}

This generation-prompt revision-regeneration loop can be repeated.
In our main experiments, we apply it once, as this already provides substantial gains; further analysis is provided in Section~\ref{sec:exp_abl}.
Within \fsname, \vsname-guided prompt redesign leverages prior failures to refine prompts and exploit favorable noise configurations. 
By treating partially correct generations as informative feedback rather than discarding them, \fsname makes more effective use of the generator’s previously expended compute and improves output fidelity.

\section{Experiments}
We comprehensively evaluate \fsname and \vsname for inference-time scaling.
First, we study the effect of prompt redesign under a fixed compute budget (Section~\ref{sec:exp_bon}).
Next, we analyze the scaling behavior of \fsname by expanding the generator’s compute budget or iteratively revising prompts, and examine its integration with visual scaling algorithms originally designed for fixed prompts (Section~\ref{sec:exp_add}).
Finally, we assess the ability of \vsname and existing verifiers to select the best-quality sample from mid-quality candidates (Section~\ref{sec:exp_verifier}) and conduct ablations on both \fsname and \vsname (Section~\ref{sec:exp_abl}).

% $k$ 
\vspace{-1em}
\begin{table*}[t]
\vspace{-0.1in}
\centering
\caption{\textbf{Quantitative comparisons of T2V generation on VBench-2.0.} $^{*}$ denotes results obtained using the standard prompt expansion, and \textbf{bold} indicates the best results. We use $N=20$ samples for Wan2.1-1.3B (small) and $N=10$ for Wan2.1-14B (large), which can lead to the smaller model achieving higher scores due to the larger number of samples.
BoN refers to ``Best-of-$N$'' selection using fixed prompts.
}\label{tab:vbench2.0}
\vspace{-10pt}
\resizebox{0.95\linewidth}{!}{
\begin{tabular}{c l ccccc c}
\toprule
Category & Method   & \Centerstack{Dynamic Spatial\\Relationship} & \Centerstack{Dynamic\\Attribute} & \Centerstack{Motion Order\\Understanding} & \Centerstack{Human\\Interaction} & \Centerstack{Composition}  & \textbf{\Centerstack{Average}} \\
\midrule

\multirow{7}{*}{\Centerstack{Controllability \\ \& Creativity }} &  Wan2.1-1.3B$^{*}$~\citep{wan2025}  & 35.56 & 46.67 & 52.87 & 74.44 & 48.33 & 51.57 \\ 
\cdashline{2-8}
\addlinespace[3pt]
& $+$BoN$^{*}$ ($N=20$)  & \textbf{43.33} & 53.33 & 51.72 & \textbf{90.00} & 50.2 & 57.73~\textcolor{blue}{$\uparrow{+6.16}$} \\
% \rowcolor{green!7} 
& \cellcolor{green!7} $+$\textbf{\fsname}$^{*}$ ($N=20$) 
& \cellcolor{green!7} \textbf{43.33} 
& \cellcolor{green!7} \textbf{73.33} 
& \cellcolor{green!7} \textbf{68.97} 
& \cellcolor{green!7} \textbf{90.00} 
& \cellcolor{green!7} \textbf{51.6} 
& \cellcolor{green!7} \textbf{65.45}~\textcolor{blue}{$\uparrow{+\textbf{13.88}}$} \\
\cline{2-8}\addlinespace[3pt]
& Wan2.1-14B$^{*}$~\citep{wan2025}  & 50.00 & 48.89 & 43.33 & 78.89 & 47.18 & 53.66 \\ 
\cdashline{2-8}
\addlinespace[3pt]

& $+$BoN$^{*}$ ($N=10$)  & 46.67 & 56.67 & 60.00 & 80.00 & 49.23 & 58.51~\textcolor{blue}{$\uparrow{+4.85}$} \\
& \cellcolor{green!7} $+$\textbf{\fsname}$^{*}$ ($N=10$) 
& \cellcolor{green!7}\textbf{60.00}  
& \cellcolor{green!7}\textbf{73.33} 
& \cellcolor{green!7}\textbf{66.67} 
& \cellcolor{green!7}\textbf{90.00} 
& \cellcolor{green!7}\textbf{54.23} 
& \cellcolor{green!7}\textbf{68.85}~\textcolor{blue}{$\uparrow{+\textbf{15.19}}$} \\ 

\midrule
Category & Method  & \Centerstack{Camera \\Motion} &  \Centerstack{Motion \\ Rationality} & \Centerstack{Mechanics} & \Centerstack{Material} & \Centerstack{Thermotics} & \textbf{\Centerstack{Average}} \\
\midrule
\multirow{7}{*}{\Centerstack{Commonsense \\ \& Physics}} & Wan2.1-1.3B$^{*}$~\citep{wan2025}  & 41.38 & 38.10 & 75.00 & 75.38 & \textbf{86.25} & 63.22  \\ 
\cdashline{2-8}
\addlinespace[3pt]
& $+$BoN$^{*}$ ($N=20$) & 37.93 & 35.71 & 84.00 & 73.91 & 81.48 & 62.61~\textcolor{red}{$\downarrow{-0.61}$}  \\
& \cellcolor{green!7}$+$\textbf{\fsname}$^{*}$ ($N=20$) 
& \cellcolor{green!7}\textbf{51.72} 
& \cellcolor{green!7}\textbf{50.00} 
& \cellcolor{green!7}\textbf{80.00} 
& \cellcolor{green!7}\textbf{78.26} 
& \cellcolor{green!7}70.37 
& \cellcolor{green!7}\textbf{66.07}~\textcolor{blue}{$\uparrow{+\textbf{3.46}}$} \\ 
\cline{2-8}
\addlinespace[3pt]
& Wan2.1-14B$^{*}$~\citep{wan2025}  & 36.67 & 40.00 & 83.33 & 77.78 & \textbf{79.49} & 63.45 \\ 
\cdashline{2-8}
\addlinespace[3pt]
& $+$BoN$^{*}$ ($N=10$) & \textbf{43.33} & 43.33 & \textbf{86.36} & 80.77 &  76.92 & 66.14~\textcolor{blue}{$\uparrow{+2.69}$} \\
& \cellcolor{green!7}$+$\textbf{\fsname}$^{*}$ ($N=10$) 
& \cellcolor{green!7}\textbf{43.33} 
& \cellcolor{green!7}\textbf{53.33} 
& \cellcolor{green!7}\textbf{86.36} 
& \cellcolor{green!7}\textbf{88.46}  
& \cellcolor{green!7}76.92 
& \cellcolor{green!7}\textbf{69.98}~\textcolor{blue}{$\uparrow{+\textbf{6.53}}$} \\
\bottomrule
\end{tabular}
}
\centering
\vspace{0.05in}
\includegraphics[width=0.98\textwidth]{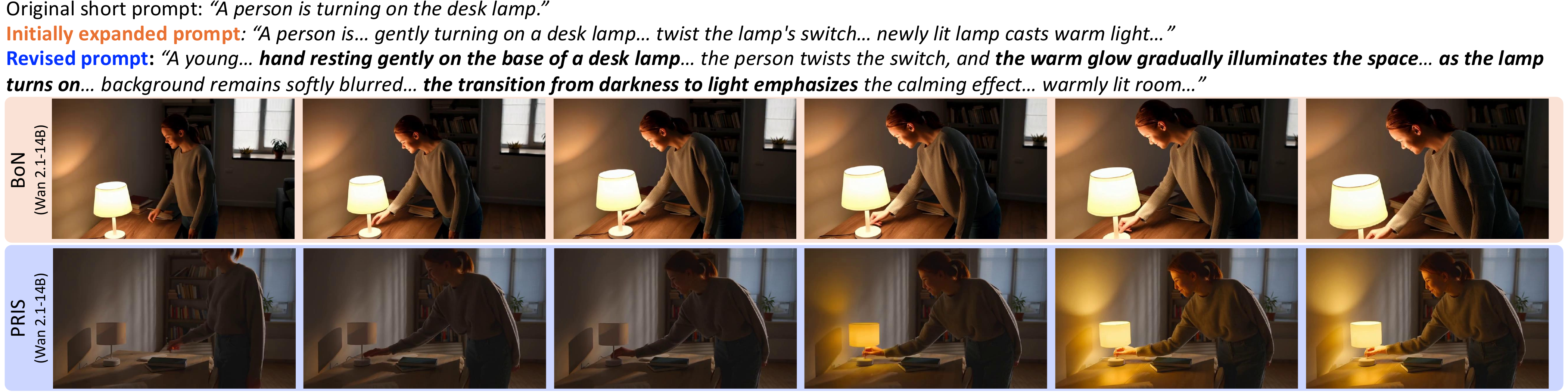}
\vspace{-0.1in}
\captionof{figure}{\textbf{Qualitative comparisons on T2V generation.}
Our revised prompt elaborates on previous failures by emphasizing causal order, ensuring the lamp turns on immediately when touched.
}\label{fig:vbench2.0}
\vspace{-0.2in}
\end{table*}
\subsection{Effect of \fsname on Inference-time Alignment and Visual Quality}\label{sec:exp_bon}
We study the effect of prompt redesign on output quality under a fixed compute budget, defined as the number of function evaluations (NFE).
For detailed experimental setups, please refer to Appendix~\ref{appen:exp_detail}.

\noindent\textbf{Experimental setup.}
For \vsname, our MLLM-based verifier, we use Qwen2.5-VL~\citep{Qwen2.5-VL} with the process-specific prompt instructions described in Section~\ref{sec:exp_verifier}, without any additional training.
We compare with Best-of-N (BoN)~\citep{ma2025scaling}, which generates $N$ samples at once and selects the best, while our method generates half of them (setting $M=\lfloor{N/2}\rfloor$), revises the prompt with feedback, and regenerates the rest using the revised prompt and top-$k$ seeds. 
We set $k=\lceil{N/4}\rceil$, thereby producing two revised prompt variants for the remaining $N-M$ samples.
We also include standard prompt expansion~\citep{wan2025} (denoted as $^{*}$) to contrast with our failure-aware revisions.
We evaluate on challenging benchmarks~\citep{li2024genaibench, zheng2025vbench2} where generations frequently exhibit partial failures.
Thus, for base generator selection, we first measure each generator’s base prompt fidelity using embedding similarity~\citep{wang2023improving} between the generated caption and the original prompt, and retain only those with sufficiently high adherence to exclude substantially weak models.

For T2I generation, we use FLUX.1-dev~\citep{labs2025flux1kontextflowmatching} on GenAI-Bench~\citep{li2024genaibench} sampling 320 prompts (20\% of the the full set) to avoid redundancy.
For the guidance reward, we utilize VQA-Score~\citep{lin2024evaluating}.
For held-out evaluation, we use DA-Score~\citep{singh2023divide} for fine-grained prompt adherence and an aesthetic predictor~\citep{laion_aesthetics_v2_2024} for image quality.
NFE is set to 2000 ($N=20$, 50 denoising steps, classifier-free guidance~\citep{ho2022classifier}, cfg=3.5).
For T2V generation, we use Wan2.1-1.3B/14B~\citep{wan2025} with VideoAlign~\citep{liu2025improving} as guidance, and evaluate on VBench2.0~\citep{zheng2025vbench2} across four dimensions: controllability, creativity, commonsense, and physical plausibility.
NFE is set to 2000 ($N=20$, 50 steps, cfg=6) for Wan2.1-1.3B and 1000 ($N=10$, 50 steps, cfg=6) for Wan2.1-14B.

\noindent\textbf{Experimental results on T2I generation.} 
We present results in Table~\ref{tab:quan_flux}, Figure~\ref{fig:qual_flux}, and Appendix~\ref{appen:more_results}. As shown in Table~\ref{tab:quan_flux}, our approach \fsname consistently outperforms all baselines across metrics. Notably, it yields substantial gains in prompt adherence while maintaining comparable aesthetic quality. Even against the standard prompt expansion variant (denoted as $^{*}$), our method achieves significantly higher scores. These results suggest that prompt expansion is most effective when guided by visual feedback, rather than by adding arbitrary details as in standard prompt expansion.
The qualitative results in Figure~\ref{fig:qual_flux} further support this claim, showing that \fsname exhibits a stronger ability to handle complex, compositional prompts compared to BoN. For the top row in Figure~\ref{fig:qual_flux}, after identifying layout specification as a challenge in the initial outputs, our method revises the prompt to emphasize layout-related details. Likewise, for the prompt ``fork not made of wood'' (bottom row in Figure~\ref{fig:qual_flux}), where BoN still produces wooden forks due to the negation, our method explicitly instructs the model to generate ``silver forks,'' thereby resolving the misunderstanding.

\noindent\textbf{Experimental results on T2V generation.} Our method delivers substantial improvements in prompt alignment for T2V generation, as shown in Table~\ref{tab:vbench2.0}, Figure~\ref{fig:vbench2.0}, and further examples in Appendix~\ref{appen:more_results}. \fsname achieves gains of $+13.88\%$ and $+15.19\%$ in the Controllability and Creativity categories for the small and large models, respectively. This significantly surpasses BoN$^{*}$, which applies standard prompt expansion at initialization without visual feedback on where to focus.
Specifically, the largest gains appear in Dynamic Attribute and Motion Order Understanding, which require sequential reasoning (e.g., ``A then B,'' ``A transitioned to B''). Here, \fsname identifies failures in the initial outputs and revises prompts to clarify how sequences should unfold, emphasizing the parts that previously failed. Qualitative examples, including revised prompt in Figure~\ref{fig:vbench2.0}, illustrate these improvements.
Beyond these categories, \fsname also improves \text{Commonsense} and \text{Physics} by $+3.46\%$ and $+6.53\%$, respectively. A notable exception is \text{Thermotics}, where performance drops slightly due to the reward model overfitting to exact numeric values rather than broader physical plausibility. Finally, while \citet{zheng2025vbench2} suggests that camera motion is largely determined by base model capacity, our results show that refining prompts to specify how camera motion should unfold in conjunction with other scene elements can still yield measurable improvements.

\begin{figure*}[t!]

\centering

\begin{minipage}[t!]{0.46\textwidth}

\centering
\includegraphics[width=0.82\linewidth]{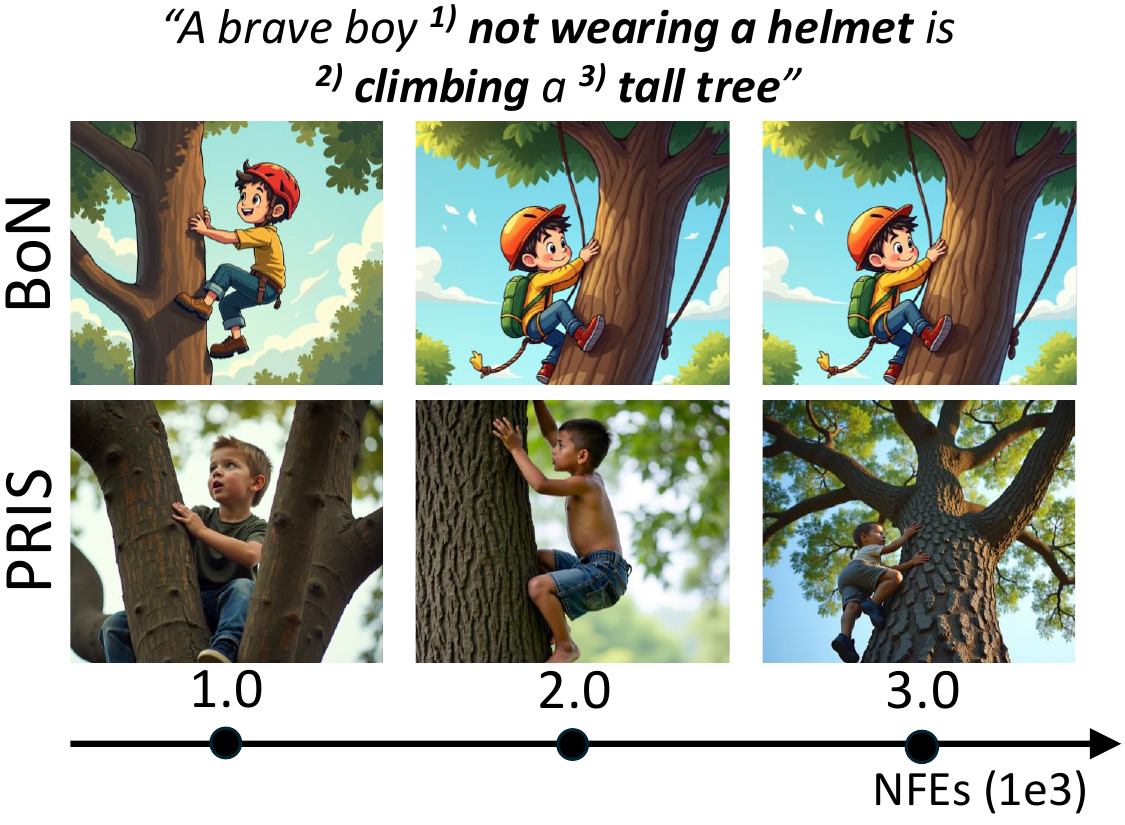}
\vspace{-0.15in}
\captionof{figure}{\textbf{Qualitative examples with increasing inference-time compute.} \fsname generates progressively taller trees while satisfying all attributes, whereas BoN consistently misses some.}
\label{fig:exp_NFE}
\end{minipage}
\hfill
\begin{minipage}[t!]{0.53\textwidth}
\vspace{-0.01in}
\captionof{table}{\textbf{Quantitative results for iterative prompt refinement with increasing inference-time compute.} 
Iteratively revision prompts consistently improves reward scores by addressing common failures, and the gains even generalize to unseen rewards. In contrast, fixed prompts often saturate and fail to transfer.
}
\centering
\label{tab:quan_iter}
\vspace{-0.1in}
\resizebox{0.98\linewidth}{!}{
\begin{tabular}{l cccc}
\toprule
\textbf{Method} & \makecell{NFEs \\(1e3)} & \makecell{VQA-\\Score \\(Given)} & \makecell{DA-Score \\ w. BLIP2-VQA \\ (Unseen)} & \makecell{Aesthetic \\Quality \\ (Unseen)} \\
\midrule
Initial (w.o. revision) & 0.5 & 0.736 & 0.679 & 5.756 \\
\midrule
Initial (w.o. revision) & 1.0 & 0.764 & 0.684 & 5.766 \\
1$^{\text{st}}$ revision & 1.0 & 0.834 & 0.703 & 5.755 \\
\midrule
Initial (w.o. revision) & 1.5 & 0.776 & 0.683 & 5.751 \\
2$^{\text{nd}}$ revision & 1.5 & 0.849 & 0.705 & 5.740 \\
\bottomrule
\end{tabular}
}
\end{minipage}
\vspace{-0.1in}
\end{figure*}

\subsection{Scaling Behaviors of \fsname}\label{sec:exp_add}
\noindent\textbf{\fsname scales with increasing NFEs.}
We provide both quantitative and qualitative evidence that \fsname scales with increasing NFEs, increasing the number of samples $N$, whereas a fixed prompt quickly saturates and fails to scale (Figures~\ref{fig:main_figure} and~\ref{fig:exp_NFE}).
In Figure~\ref{fig:main_figure}, BoN, which relies on fixed prompts, plateaus on held-out evaluation beyond 1e3 NFEs, while \fsname continues to improve, both quantitatively in reward score and qualitatively in visual alignment.
Similarly, Figure~\ref{fig:exp_NFE} further illustrates this trend qualitatively: \fsname produces progressively taller trees while satisfying all prompt elements, even at smaller budgets, whereas BoN repeatedly fails, generating a boy wearing a helmet.

\noindent\textbf{Effectiveness of iterative prompt revisions with \fsname.}
Table~\ref{tab:quan_iter} evaluates whether iterative revisions, which update prompts based on newly identified failures, provide benefits beyond a single iteration of revision. As shown, iterative revision yields consistent improvements across both given and held-out metrics for prompt adherence while maintaining comparable aesthetic quality. 
The first update brings a substantial gain, and the second adds further improvements, suggesting that iterative revision progressively strengthens alignment.
While multiple revisions yield cumulative gains, the first update already offers a substantial improvement; therefore, we adopt a single refinement step in the main experiments.
Moreover, such gains do not appear without \fsname. Simply generating more samples with increased compute budget leads to saturated performance on unseen rewards. This highlights that targeted prompt correction is more effective than brute-force visual scaling.

\begin{figure*}[t!]
\centering

\begin{minipage}[t]{0.5\textwidth}
\vspace{-0.01in}
\centering
    \captionof{table}{\textbf{Quantitative results of integrating \fsname with T2I visual scaling methods} on GenAI-Bench. BoN refers to ``Best-of-N'' selection using fixed prompts. \textbf{Bold} shows the best.}
\label{tab:quan_combine}
\vspace{-0.1in}
\resizebox{0.92\linewidth}{!}{
\begin{tabular}{l ccc}
\toprule
\textbf{Method} & \makecell{VQA-\\Score\\(Given)} & \makecell{DA-Score \\ w. BLIP2-VQA\\(Unseen)} & \makecell{Asthetic \\ Quality \\(Unseen)} \\
\midrule
SDXL~\cite{podell2023sdxl} & 0.639 & 0.652 & 5.759 \\
\cdashline{1-4}
\addlinespace[3pt]
$+$BoN &  0.649 & 0.663 & 5.810 \\
$+$DAS~\cite{kim2025testtime} & 0.657 & 0.671 & 5.819 \\
\rowcolor{green!7}
$+$DAS w/ \textbf{\fsname} & \textbf{0.700} & \textbf{0.688} & \textbf{5.897} \\
\midrule
FLUX.1-schnell~\cite{flux2024} & 0.672 & 0.676 & 5.519 \\
\cdashline{1-4}
\addlinespace[3pt]
$+$BoN & 0.869 & 0.704 & 5.497 \\
$+$RBF~\cite{kim2025inference} & 0.922 & 0.706 & 5.426\\
\rowcolor{green!7}
$+$RBF w/ \textbf{\fsname} & \textbf{0.936} & \textbf{0.723} & \textbf{5.528}\\
\bottomrule
\end{tabular}
}
\end{minipage}
\hfill
\begin{minipage}[t]{0.47\textwidth}
\vspace{-0.03in}
\centering
    \includegraphics[width=\linewidth]{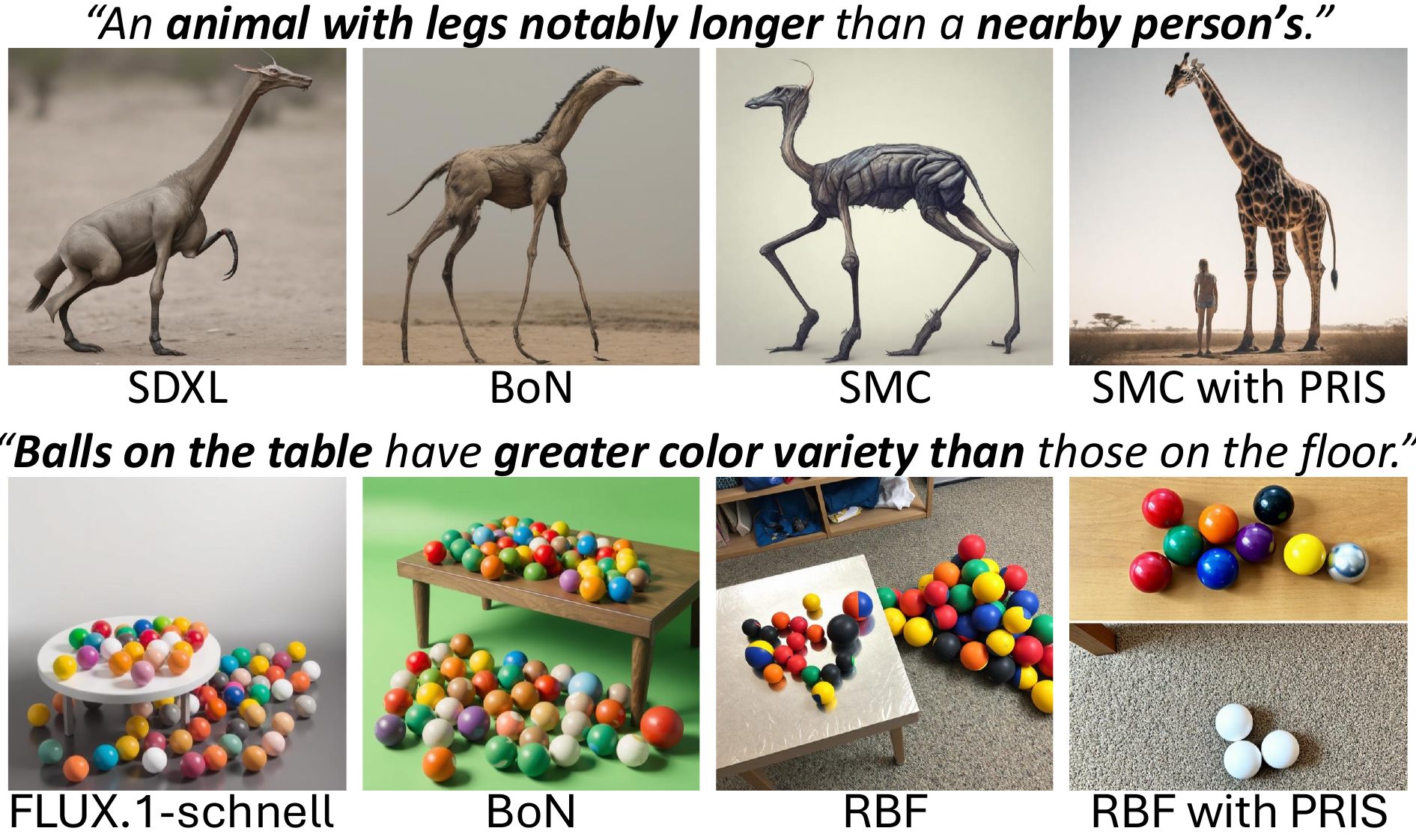}
\vspace{-0.26in}
    \caption{\textbf{Qualitative examples of integrating \fsname with T2I noise-scaling baselines.}}
    \label{fig:qual_combine}
\end{minipage}

\vspace{-0.1in}
\end{figure*}

\noindent\textbf{Integration with visual scaling methods beyond BoN.}
\fsname is complementary to existing visual scaling methods that expand the sampling space with fixed prompts.
While these approaches modify noise or sampling dynamics, \fsname targets prompt-level failures and can be combined with them by enabling prompt revision.
To validate this complementarity, we integrate \fsname with two established T2I methods, DAS~\citep{kim2025testtime} and RBF~\citep{kim2025inference}, following their original protocols and evaluating on GenAI-Bench consistent with our main setup. Table~\ref{tab:quan_combine} and Figure~\ref{fig:qual_combine} show that integration yields superior alignment on both given and unseen rewards. Notably, whereas RBF often sacrifices aesthetics to improve alignment, our approach improves both simultaneously. Qualitatively, Figure~\ref{fig:qual_combine} further shows that although DAS and RBF alone struggle on difficult prompts, their integration with \fsname resolves these cases, producing outputs that are both prompt-aligned and visually coherent. Full experimental details, additional examples, and T2V results are provided in Appendix~\ref{appen:more_analysis}.\\

\vspace{-2em}
   
\subsection{Evaluating the Verification Accuracy of \vsname}\label{sec:exp_verifier}
We introduce \vsname to address the lack of fine-grained evaluation in text-visual alignment, enabling element-level verification for precise and interpretable assessment. As prompts become more complex, it is critical to check whether all attributes are satisfied rather than relying on a single opaque score. However, widely used human preference datasets~\citep{liu2025improving, liu2025videodpo, LiFT, he2024videoscore, xu2024visionreward} are insufficient as they provide only pairwise judgments and do not capture whether a single video fully satisfies the prompt. Moreover, they do not reflect inference-time needs, where a verifier must pick the best-aligned sample from a diverse pool of mid-quality outputs. To fill this gap, we construct a benchmark that pairs prompts with both fully aligned (ground-truth) and partially aligned (distractors) visuals, covering varying degrees of completeness and fidelity. Additional dataset details and analyses are provided in Appendix~\ref{appen:benchmark}.

\noindent\textbf{Constructing the benchmark.}
Each prompt is paired with multiple aligned and partially aligned videos, along with tags indicating reasons for misalignment for the partially aligned cases. Prompts are drawn from widely used video model demos and from VBench 2.0~\citep{zheng2025vbench2}, yielding a total of 410 prompts. Candidate videos are generated using diverse state-of-the-art closed- and open-source models~\citep{veo2025, kling2025, wan2025}. Several human annotators mark a video as aligned if it fully satisfies the prompt and provide explanations when they label it as misaligned. The final label for each video is determined by majority vote.

\noindent\textbf{Evaluation setup and baselines.}
We evaluate \vsname and existing verifiers on our benchmark which simulates the inference-time scaling setting. As baselines, we consider widely used learned reward models~\citep{liu2025improving, xu2024visionreward, UnifiedReward}, trained on preference datasets to output a scalar score per video. We then evaluate \vsname itself, which performs zero-shot prompt-adherence verification using MLLMs~\citep{Qwen2.5-VL}, along with an ablation that removes its factual correction component. In this ablation, verification is reduced to decomposed visual QA, where each element is judged independently via QA, and the final score is determined by the number of elements marked aligned. For all methods, we select the top-scoring video and evaluate against human annotations.

\begin{table}[h]
\vspace{-0.1in}
\centering
\caption{\textbf{Quantitative results on verifier accuracy} in selecting GT visual outputs. \textbf{Bold} indicates the best results.}
\label{tab:quan_dataset}
\vspace{-0.1in}
\resizebox{0.6\linewidth}{!}{
\begin{tabular}{l c}
\toprule
Verifier & Accuracy \\
\midrule
VisionReward~\citep{xu2024visionreward} & 0.571 \\ 
UnifiedReward~\citep{UnifiedReward} & 0.498 \\ 
VideoAlign~\citep{liu2025improving} & 0.693 \\ 
\midrule
% Binary VQA & -- \\
Decomposed binary VQA & 0.700 \\
\fsname (Ours) & \textbf{0.763} \\
\bottomrule
\end{tabular}
}
% \vspace{-0.2in}
\end{table}

\noindent\textbf{Evaluation results.} Table~\ref{tab:quan_dataset} shows that \vsname achieves the highest accuracy, significantly surpassing even VideoAlign, the strongest reward model. 
Moreover, unlike learned reward models, \vsname provides fine-grained, interpretable reasoning even without training. It also outperforms decomposed VQA, supporting our design choice of factual correction with text-based verification, consistent with recent findings that text-based measures are more reliable than direct VQA~\citep{fu2025hidden, bai2024hallucination}.

\subsection{Ablation Study and Analysis}\label{sec:exp_abl}
\noindent\textbf{Impact of common-failure-aware prompt redesign.} We conduct an ablation comparing our redesign strategy, which revises prompts based on failure patterns consistently shared across the top-$k$ best-aligned samples, against a per-sample prompt revision baseline that attempts to fix failures independently for each sample.
Our method focuses revisions on attributes that are systematically difficult for the model to generate, rather than dispersing corrections across noisy, sample-specific failures.
As shown in Table~\ref{tab:abl}, common-failure revision consistently outperforms per-sample revision in both T2I and T2V.

\begin{table}[h]
\vspace{-0.06in}
\captionof{table}{\textbf{Ablation study of \fsname.} \# \textit{d.e.} and \# \textit{c.f.} denote the numbers of decomposed elements and common failures, respectively; \textbf{bold} indicates the best.}
\vspace{-0.01in}
\centering
\label{tab:abl}
\vspace{-0.09in}
\resizebox{0.98\linewidth}{!}{
\begin{tabular}{c l c c c}
\toprule
Task & Prompt Revision &  Avg. \# \textit{d.e.} & Avg. \# \textit{c.f.} & Score \\
\midrule

\multirow{3.5}{*}{T2I} & 
 w/o revision &  &  & 0.783 \\
\cdashline{2-5}
\addlinespace[3pt]

    & Per-sample & \multirow{2}{*}{3.5} & - & 0.853~\textcolor{blue}{$\uparrow{+\text{0.070}}$}\\
    %{\scriptsize$\uparrow{+\textbf{6.53}}$}\\
    & Common-failure &  & 0.72 & \textbf{0.854}~\textcolor{blue}{$\uparrow{+\textbf{0.071}}$} \\
\midrule
\multirow{3.5}{*}{T2V} 
& w/o revision &  &  & 0.711 \\
\cdashline{2-5}
\addlinespace[3pt]

    & Per-sample & \multirow{2}{*}{7.3} & - & 0.619~\textcolor{red}{$\downarrow{-\text{0.092}}$} \\
    & Common-failure & & 1.46 & \textbf{0.754}~\textcolor{blue}{$\uparrow{+\textbf{0.043}}$} \\
\bottomrule
\end{tabular}
}
\vspace{-0.05in}
\end{table}

Notably, in T2V generation, where fully aligning the generated visuals with the prompt is considerably more challenging than in T2I, per-sample revision performs worse than applying no revision at all. Attempting to correct every observed error dilutes the update signal, spreads compute on low-probability or seed-specific artifacts, and ultimately fails to address the truly persistent misalignments.
In contrast, our method anchors prompt updates to high-likelihood failures that repeatedly surface across strong samples, making revisions both focused and efficient.
Finally, the number of common failures (\textit{c.f.} in Table~\ref{tab:abl}) observed among top-performing seeds empirically supports our key motivation: different seeds do share recurring failure modes, and exploiting these shared patterns becomes increasingly essential for effective prompt redesign as prompt complexity grows.

\noindent\textbf{Compute time analysis.}
In our experiments, we follow the standard practice of comparing methods under the same NFEs~\citep{ma2025scaling, kim2025inference}. We also evaluate under matched total wall-clock time in Table~\ref{tab:abl_time}, including verifiers. 
Even under this setting, allocating wall-clock time to our framework is more effective than simply increasing NFEs (i.e., increasing $N$) for the generator. 
Although \vsname introduces a modest overhead, mainly from captioning, \fsname achieves substantially larger gains in prompt adherence. These results indicate that directing wall-clock time toward verifier-guided prompt revision is more beneficial than spending the same time on brute-force generation. 
Please refer to Appendix~\ref{appen:more_analysis} for a detailed breakdown of the compute time for verification.
\begin{table}[h]
\vspace{-0.1in}
\centering
\caption{\textbf{Quantitative evaluation with matched compute.}}
\label{tab:abl_time}
\vspace{-0.12in}
\resizebox{0.6\linewidth}{!}{
\begin{tabular}{c l c c}
\toprule
Task & Method & NFEs (1e3) & Score \\
\midrule
\multirow{2}{*}{T2I} 
    & BoN & 4.0 & 0.790 \\ %N=40 
    & \fsname & 1.0 & \textbf{0.834} \\ %N=10
\midrule
\multirow{2}{*}{T2V}
    & BoN & 4.0 & 0.935 \\ %N=40
    & \fsname & 2.0 & \textbf{0.964} \\ %N=20
\bottomrule
\end{tabular}
}
\vspace{-0.24in}
\end{table}
\section{Conclusion}
We address an overlooked gap in inference-time scaling by redesigning prompts from common failures in samples, rather than relying on visual-only scaling or isolated per-sample prompt updates as in prior work.
We introduce \vsname, a verifier that provides fine-grained text-visual alignment assessments, and \fsname, a \vsname-guided framework that revises prompts based on common failure patterns observed across generated outputs.
By reviewing generated visuals to identify recurring misalignments, \fsname adaptively updates the prompt in response to the generator’s failures.
Across both T2I and T2V settings, our approach achieves substantial gains in prompt adherence, demonstrating that jointly scaling prompts and visuals yields stronger scaling behavior than visual-only methods.
Furthermore, our findings show that leveraging recurring failure patterns for prompt redesign is crucial to fully realize the benefits of inference-time scaling.

{
    \small
    \bibliographystyle{ieeenat_fullname}
    \bibliography{main}
}

% \appendix
% \clearpage
% \input{appendix/more_analysis}
% \input{appendix/benchmark}
% \input{appendix/exp_detail}
% \input{appendix/more_results}

% \onecolumn
% \begin{center}
% {\bf {\Large Appendix: Rethinking Prompt Design \\for Inference-time Scaling in Text-to-Visual Generation}} 
% \end{center}

\clearpage
\appendix
\onecolumn
\setcounter{page}{1}
\begin{center}
{
\Large
\textbf{Rethinking Prompt Design for Inference-time Scaling in Text-to-Visual Generation}\\
\vspace{0.5em}Supplementary Material \\
}
\end{center}

\section{Additional Analysis}\label{appen:more_analysis}

\subsection{Qualitative Examples of Common Failures}\label{appen:failure_pattern}
\begin{figure}[h]
    \centering
    \vspace{-0.15in}
    \includegraphics[width=0.7\linewidth]{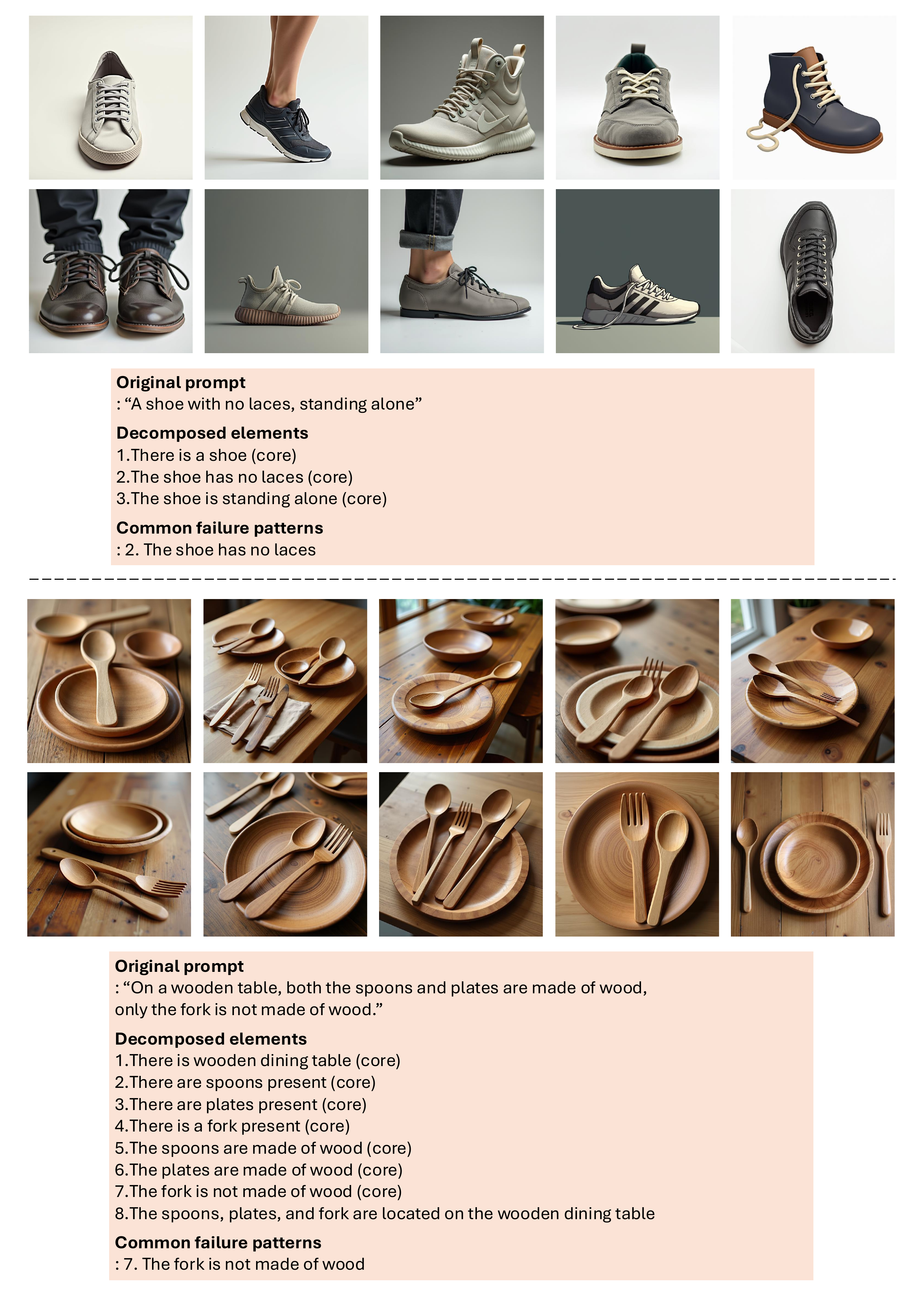}
    \vspace{-0.1in}
    \caption{\textbf{Qualitative examples of recurring misalignments} when generating multiple images from a fixed prompt, with decomposed elements and common failures identified by \vsname.}
    \label{fig:appen_failure_t2i}
    \vspace{-0.2in}
\end{figure}

\begin{figure}[ht]
    \centering
    \vspace{-0.1in}
    \includegraphics[width=0.8\linewidth]{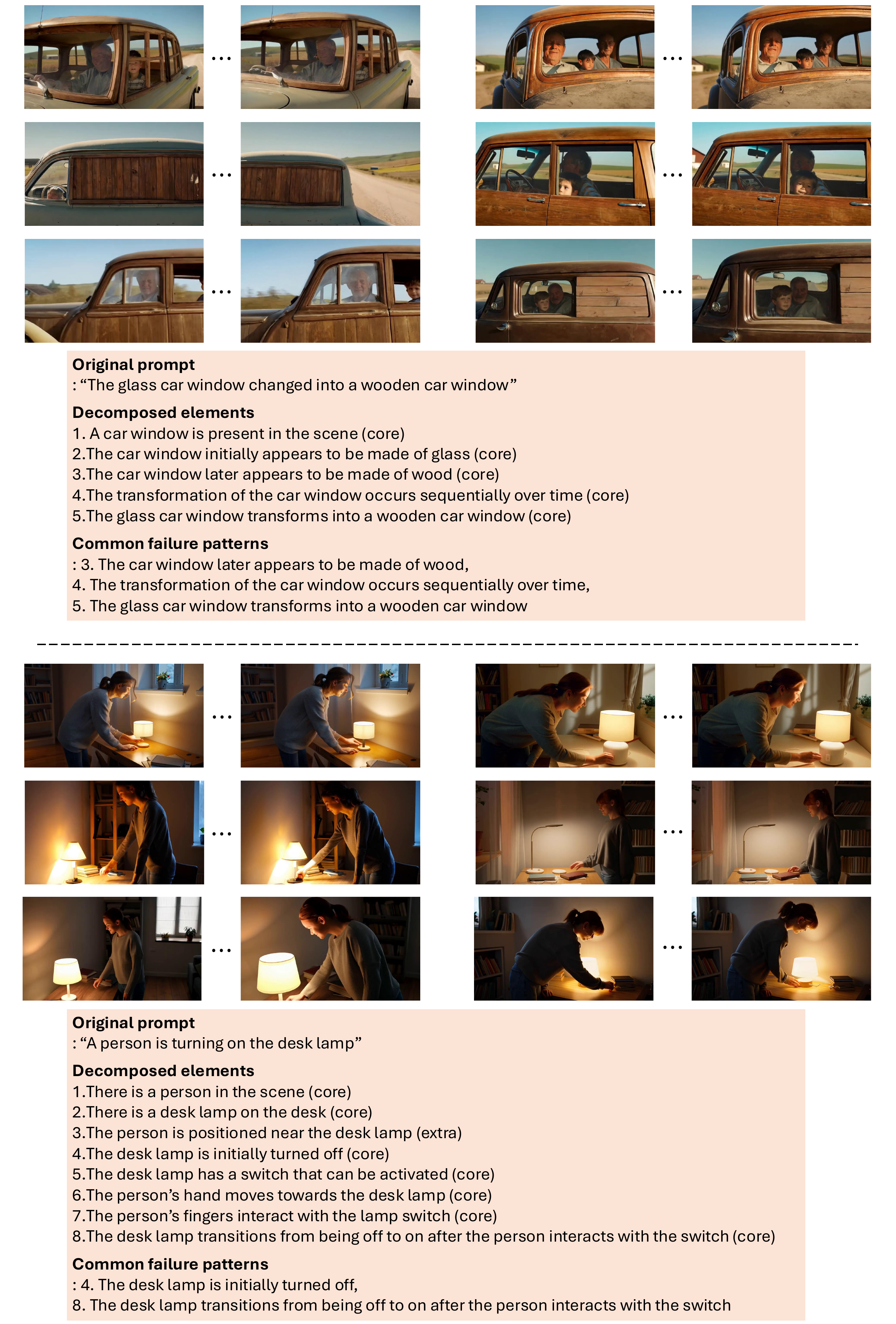}
    \caption{\textbf{Qualitative examples of recurring misalignments} when generating multiple videos from a fixed prompt, with decomposed elements and common failures identified by \vsname. We illustrate the first and the last frame for each generated video.}
    \label{fig:appen_failure_t2v}
    \vspace{-0.2in}
\end{figure}

We present qualitative examples of the identified common failure patterns for text-to-image generation in Figure~\ref{fig:appen_failure_t2i} and for text-to-video generation in Figure~\ref{fig:appen_failure_t2v}.

\clearpage

\subsection{Details of \vlname (\vsname)}
\begin{figure}[ht]
    \centering
    \vspace{-0.1in}
    \includegraphics[width=\linewidth]{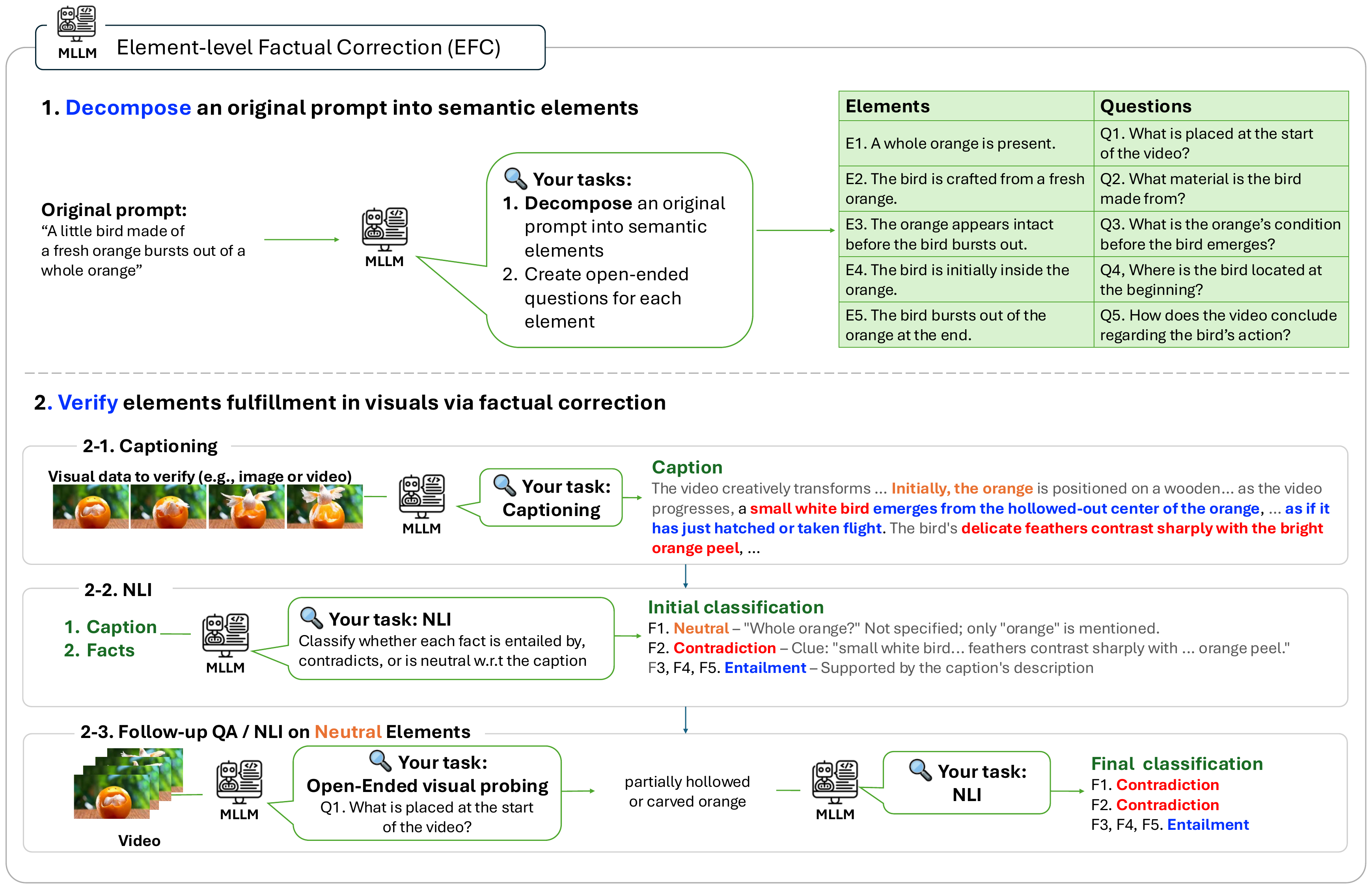}
    \caption{\textbf{Illustration of \vsname.} The figure illustrates how \vsname provides fine-grained, interpretable verification of prompt adherence. It first decomposes the prompt into semantic elements, then generates captions from the visuals, and applies factual correction to classify each element as entailment, neutral, or contradiction. Elements initially labeled neutral (due to missing mentions in the caption) are reevaluated to decide between entailment and contradiction. This design avoids direct QA, leading to more accurate verification.}
    \label{fig:appen_verifier}
    \vspace{-0.2in}
\end{figure}

 We present a detailed overview of the visual verification process in our verifier, \vlname (\vsname) in Figure~\ref{fig:appen_verifier}.

\noindent\textbf{Element-level factual correction.} The goal of this process is to provide fine-grained and interpretable feedback on whether each part of a prompt is faithfully realized in the generated visuals.
Given a prompt and its corresponding outputs (images or videos), \vsname first decomposes the prompt into multiple disjoint semantic elements using a system prompt. For each element, it also constructs an open-ended question, where the element itself serves as the expected answer.
Next, \vsname verifies the fulfillment of these elements in the generated visuals through factual correction. Instead of relying on visual question answering, our key idea is to perform text-based comparison between the semantic elements and the visuals. 
To enable this, \vsname first extracts captions from the generated visuals and then applies natural language inference (NLI) to classify each element as entailment, neutral, or contradiction. 

\noindent\textbf{Open-Ended visual probing.} For elements classified as neutral, where captions are missing or ambiguous, \vsname reuses the previously generated open-ended questions, queries the visual input again, and applies a second NLI step to the corresponding free-form answers, assigning a final label of either entailment or contradiction.
Unlike direct QA, this procedure asks open-ended questions and compares their answers with the target elements to determine whether the expected element is present, rather than relying on yes/no responses. This removes affirmation bias inherent in binary QA and avoids providing contextual cues that may cause the verifier to rely on textual hints instead of extracting information directly from the visuals.
Through this process, \vsname pinpoints which parts of the prompt are faithfully represented and which are contradicted, thereby enabling accurate and interpretable fine-grained feedback for generated visuals.

\clearpage
\subsection{Details on Integration Beyond BoN}
This section provides additional details on the integration of our framework with visual scaling methods, complementing Section~\ref{sec:exp_add}.

\begin{figure}[ht]
    \centering
    \vspace{-0.1in}
    \includegraphics[width=0.9\linewidth]{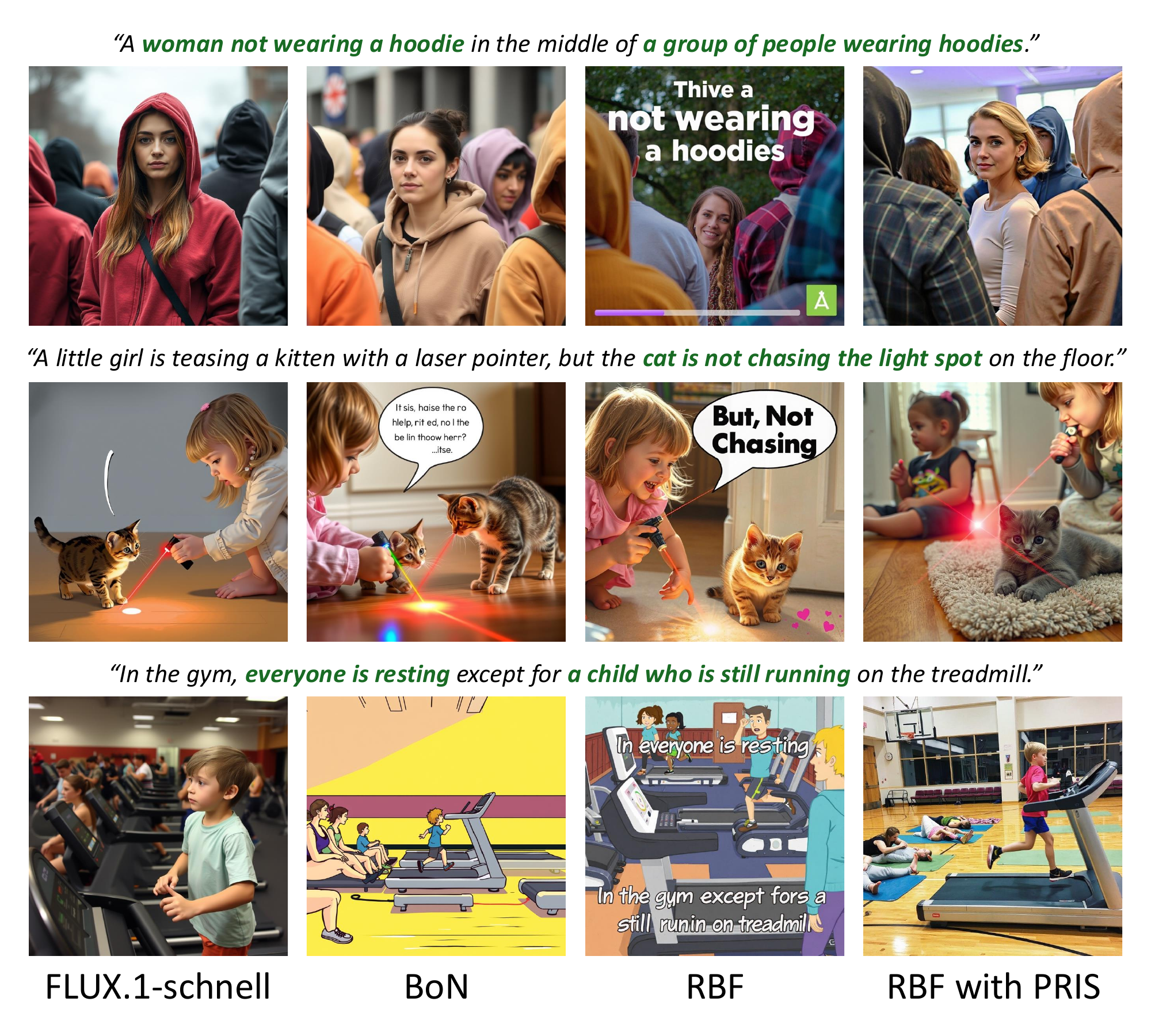}
    \vspace{-0.2in}
    \caption{\textbf{Qualitative artifact results with RBF.} RBF alone often generates visuals where the prompt text is directly rendered on the image due to reward over-optimization, whereas combining RBF with our method substantially alleviates this issue.}
    \label{fig:appen_rbf_artifact}
    \vspace{-0.05in}
\end{figure}

\noindent\textbf{Experiments on text-to-image generations.}
We integrate our approach with two inference-time scaling methods focused on visuals: DAS~\citep{kim2025testtime} and RBF~\citep{kim2025testtime}. Following their original experimental protocols, we use SDXL~\citep{podell2023sdxl} for DAS and Flux.1-schnell~\citep{labs2025flux1kontextflowmatching} for RBF. In both settings, we generate a total of 8 samples, divided into two batches of 4. When combined with our method, the first batch of 4 samples is generated, the prompt is revised, and another 4 samples are generated, ensuring that the total number of function evaluations remains equivalent.

\begin{figure}[t]
    \centering
    \vspace{-0.2in}
    \includegraphics[width=0.75\linewidth]{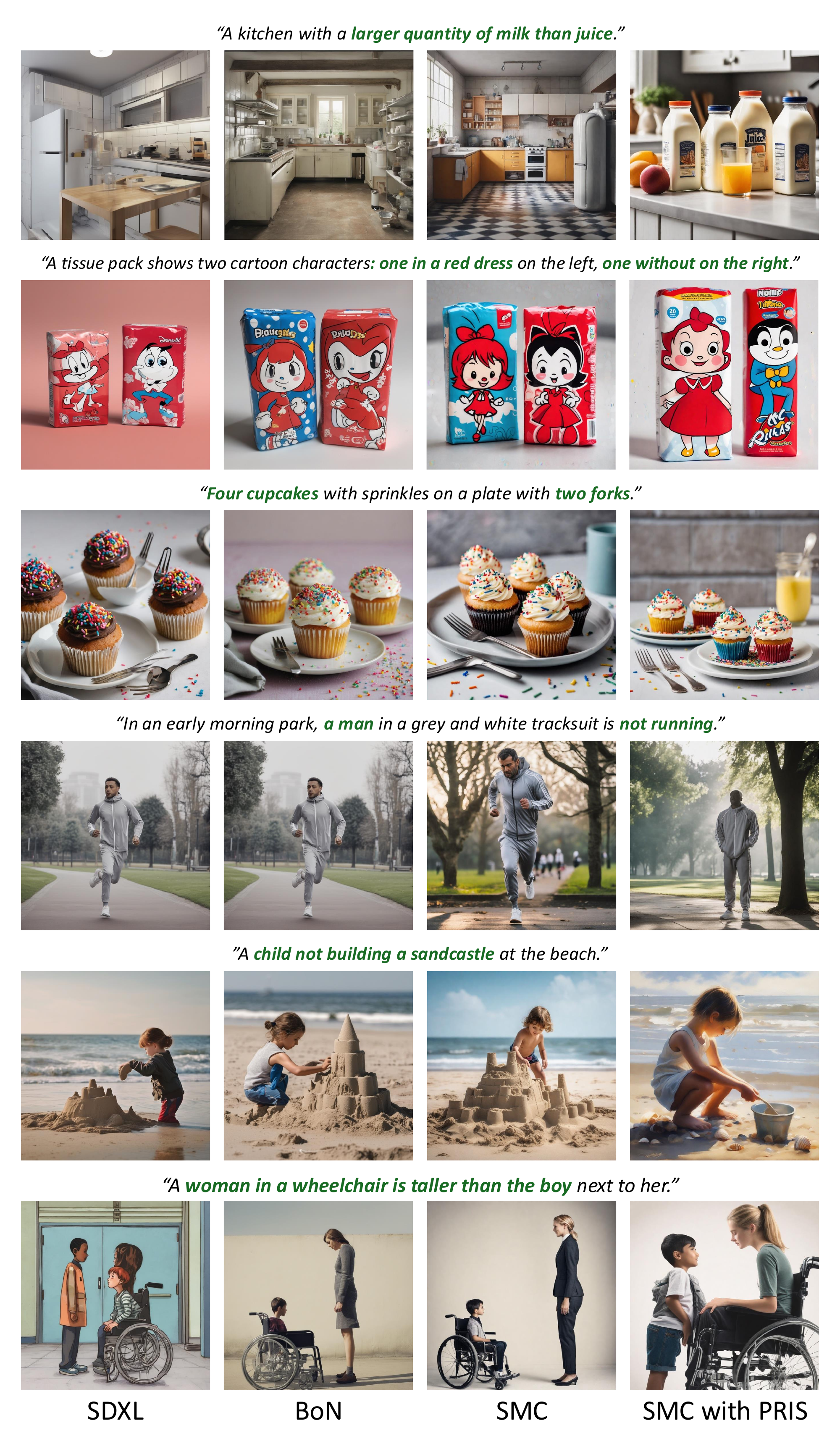}
    \vspace{-0.2in}
    \caption{\textbf{Qualitative comparisons when integrating our method with SMC} under the same compute budget. Our approach more faithfully follows the prompt, effectively enabling SMC to scale visuals.}
    \label{fig:appen_smc}
    \vspace{-0.2in}
\end{figure}

\begin{figure}[t]
    \centering
    \vspace{-0.2in}
    \includegraphics[width=0.75\linewidth]{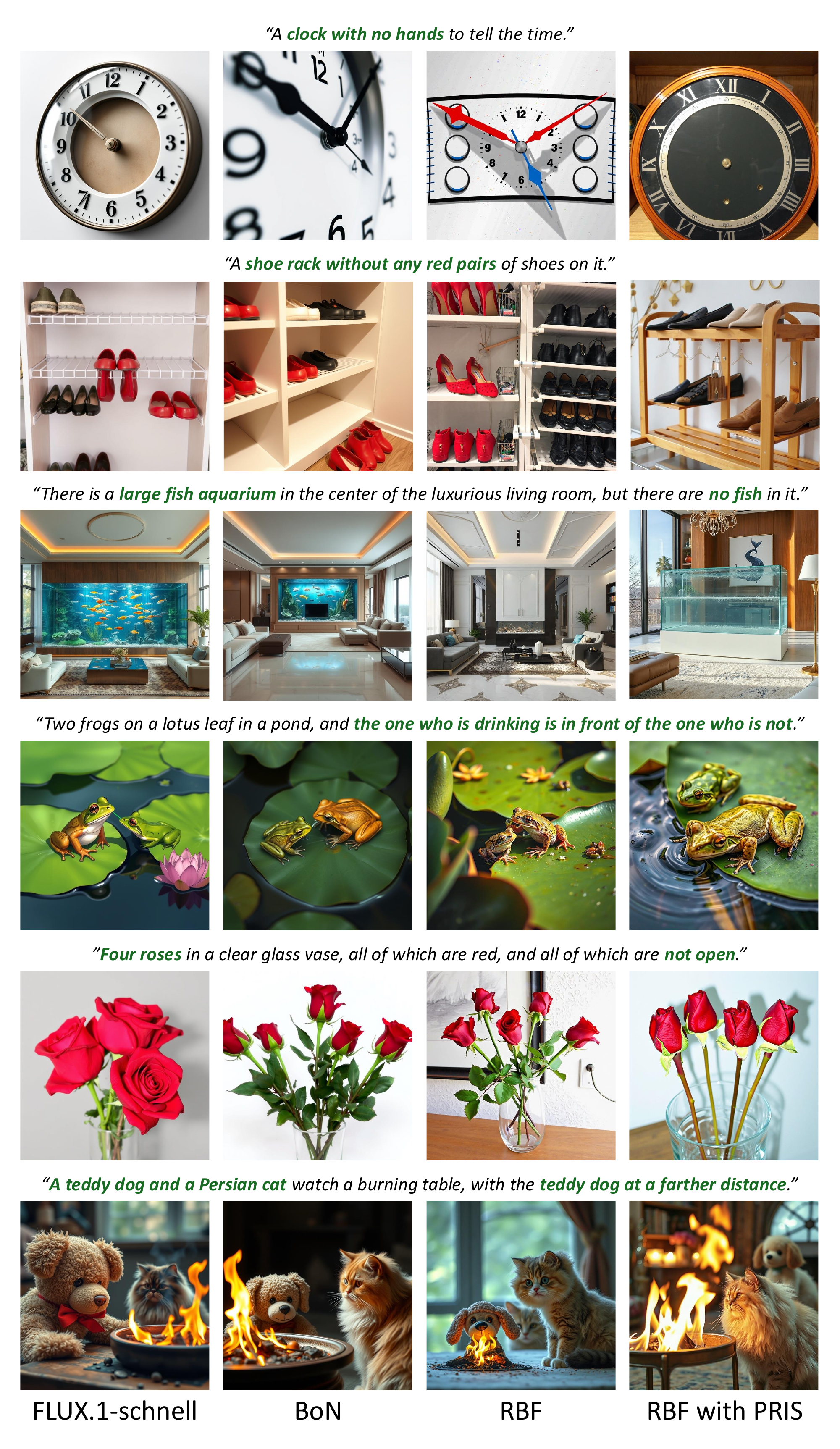}
    \vspace{-0.2in}
    \caption{\textbf{Qualitative comparisons of RBF integrated with our method} under the same compute budget. Our method adheres more closely to the prompt and further improves RBF’s visual scaling.}
    \label{fig:appen_rbf}
    \vspace{-0.2in}
\end{figure}

In addition to Table~\ref{tab:quan_combine} and Figure~\ref{tab:quan_combine} in the main manuscript, Figures~\ref{fig:appen_smc} and \ref{fig:appen_rbf} demonstrate that our integrated results achieve substantially better prompt adherence than visual scaling alone, for DAS and RBF, respectively. This indicates that advanced visual scaling methods can be further enhanced when combined with scaled prompts.
It is also noteworthy that scaling visuals alone often leads to undesired outcomes caused by reward over-optimization (see Figure~\ref{fig:appen_rbf_artifact}). In such cases, the model may even render the textual prompt itself, since these images achieve artificially high reward scores. For example, Figure~\ref{fig:appen_rbf_artifact} shows that RBF frequently generates images where the prompt text is printed directly. By contrast, our method mitigates this issue: the revised prompt guides the generator, while the original prompt is used only for the reward signal. This separation effectively reduces over-optimization artifacts and yields more faithful generations, even when \fsname is combined with RBF.

\clearpage
\noindent\textbf{Experiments on text-to-video generations.}
We integrate our approach with EvoSearch~\citep{he2025scaling}, following its original setup on Wan2.1-1.3B. EvoSearch uses an evolution schedule of [5, 20, 30, 45] and a population schedule of [10, 5, 5, 5], totaling 2,000 NFEs. For integration, we first generate 10 samples with 50 steps (1,000 NFEs), then allocate the remaining 940 NFEs with [5, 30] for the evolution schedule and [5, 4] for the population schedule, resulting in 60 fewer NFEs than EvoSearch. As in the main manuscript, we evaluate on VBench2.0 with VideoAlign as the guiding reward.

\begin{figure}[ht]
    \centering
    \vspace{-0.2in}
    \includegraphics[width=\linewidth]{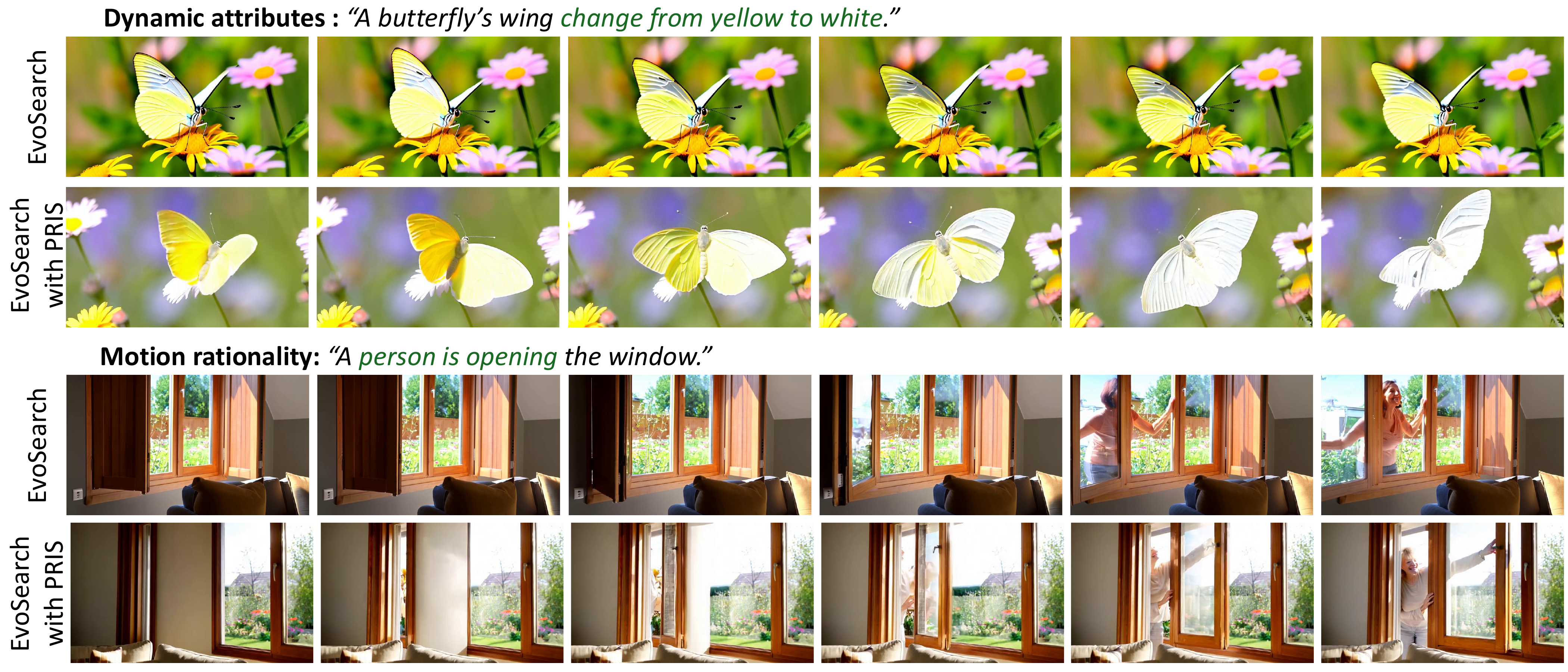}
    \vspace{-0.3in}
    \caption{\textbf{Qualitative examples comparing EvoSearch and EvoSearch+PRIS}. In the first case, EvoSearch fails to change the butterfly’s wing color despite scaling, whereas our method succeeds. In the second case, EvoSearch depicts the window as already open before the person attempts to open it, while our method correctly shows the window opening as the person reaches out.}
    \label{fig:appen_evo}
    \vspace{-0.1in}
\end{figure}

Table~\ref{tab:appen_t2v_combine} and Figure~\ref{fig:appen_evo} present the quantitative and qualitative results. Unlike EvoSearch, which was evaluated on relatively simple prompts, our experiments employ more complex ones. In this setting, EvoSearch scores degrade after scaling, suggesting limited generalization to the unseen reward of VBench2.0. By contrast, when integrated with our method, it achieves improved average scores on VBench2.0.

\begin{figure*}[ht]
\centering
\vspace{-0.05in}
\captionof{table}{\textbf{Quantitative T2V results on VBench2.0, comparing EvoSearch alone with EvoSearch integrated with PRIS}. EvoSearch fails to generalize to unseen rewards, whereas integration with PRIS improves performance.}
\label{tab:appen_t2v_combine}
\vspace{-0.12in}
\resizebox{0.75\linewidth}{!}{
\begin{tabular}{l c c c c}
\toprule
Method   & \Centerstack{Motion\\Rationality} & \Centerstack{Motion Order\\Understanding} & \Centerstack{Dynamic\\Attribute} & \textbf{\Centerstack{Average}} \\
\midrule
Wan2.1-1.3B & 38.10 & \textbf{52.87} & 46.67 & 45.88 \\
\cdashline{1-5}
\addlinespace[3pt]

EvoSearch & 32.14 & 51.72 & 43.33 & 42.20~\textcolor{red}{\scriptsize$\downarrow{-3.68}$} \\ 
EvoSearch + PRIS & \textbf{53.57} & 48.28 & \textbf{60.00} & \textbf{53.95}~\textcolor{blue}{$\uparrow{+8.07}$} \\

\bottomrule
\end{tabular}
}
\end{figure*}

\vspace{-0.05in}

\subsection{Detailed Computational Time Analysis}
In this section, we provide a detailed breakdown of verification and generation time, complementing Section~\ref{sec:exp_abl}. All measurements are conducted on a single NVIDIA H100 80GB GPU.
For images, generating a single sample resolution $(1024, 1024)$ with Flux.1-dev takes on average 13 seconds, while verification with our verifier, \vsname, requires 41 seconds. This implies that each verification is computationally equivalent to generating approximately three additional images. 
To balance this overhead, we set the number of function evaluations (NFE) to 4000 for BoN and 1000 for our method, corresponding to 40 and 10 images, respectively (with 50 sampling steps and classifier-free guidance).
For videos, generating an 81-frame sequence at resolution $(480, 832)$ with Wan2.1-1.3B requires 105 seconds on average, while verification takes 100 seconds, approximately equivalent to one additional video generation. Accordingly, we set the NFE to 4000 for BoN (40 videos) and 2000 for our method (20 videos), again under 50 sampling steps with classifier-free guidance.

Our verifier is intentionally built on a pretrained MLLM without task-specific optimization, demonstrating that strong results can be achieved without additional training. Nonetheless, fine-tuning the base MLLM remains a promising direction for reducing verification time and improving efficiency.

\subsection{Comparison with ReflectionFlow}\label{appen:add_comp}
\begin{figure}[ht]
    \centering
    \vspace{-0.1in}
    \includegraphics[width=0.62\linewidth]{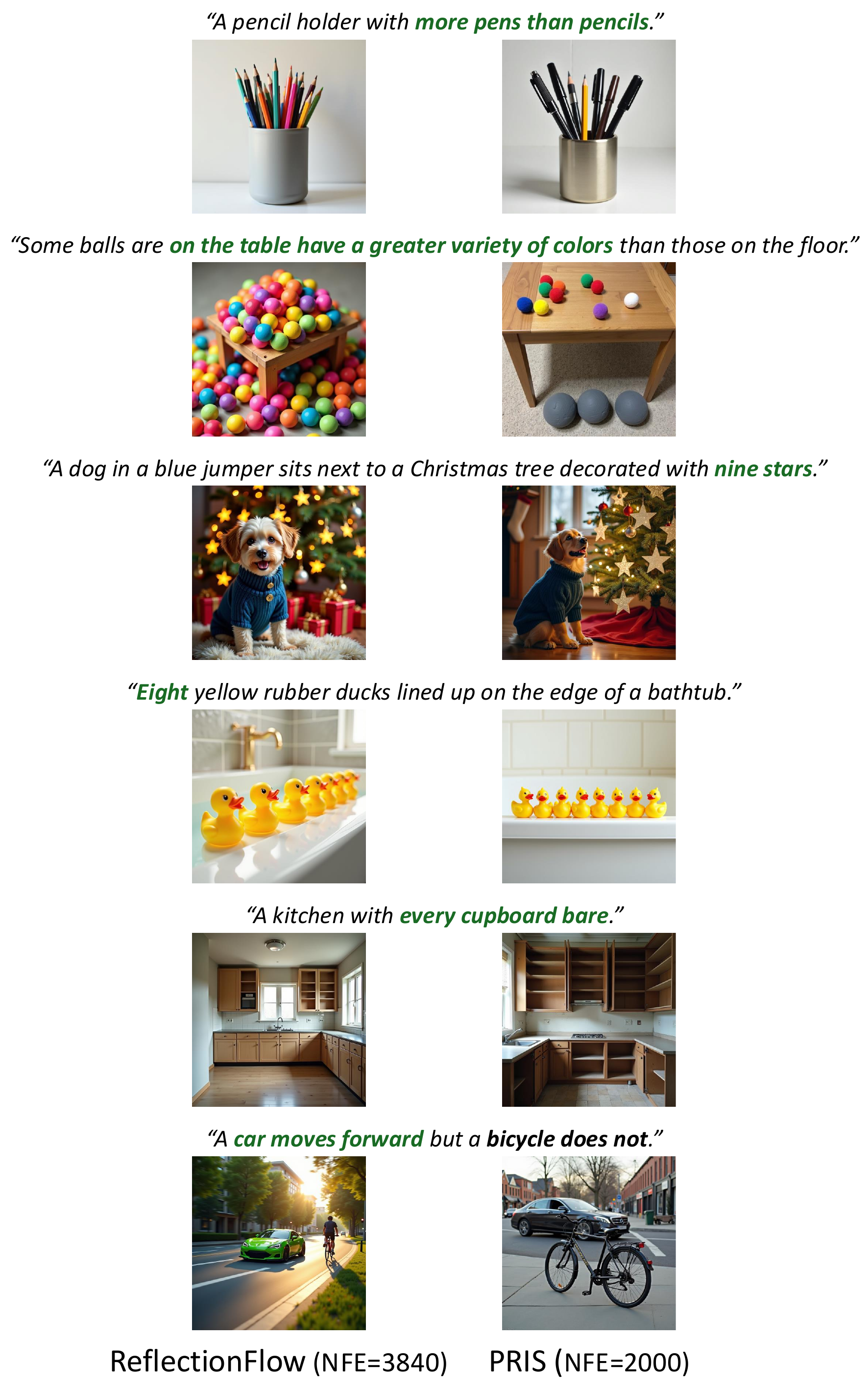}
    \vspace{-0.1in}
    \caption{\textbf{Qualitative comparisons with ReflectionFlow.} Despite being training-free, our method markedly outperforms the learned approach, underscoring the effectiveness of \textit{correcting prompts using shared failure patterns across samples}.}
    \label{fig:appen_comp_reflection}
    \vspace{-0.2in}
\end{figure}

We compare our method with ReflectionFlow~\citep{zhuo2025reflection}, which relies on a trained reflection model to iteratively edit each generated sample.
Our approach differs in three fundamental ways.
First, we revise the prompt itself based on common failure patterns across samples, rather than reacting to individual errors.
Second, our method is entirely training-free and does not rely on any auxiliary editing models.
Third, it is applicable to any text-conditioned generator, whereas ReflectionFlow requires model-specific training.
For a favorable comparison, we allocate 3840 NFEs ($N=64$) to ReflectionFlow, following its default configuration, while ours uses only 2000 NFEs ($N=20$).
Even under this compute-advantaged setup for ReflectionFlow, our method consistently produces more accurate and semantically aligned results, as shown in Figure~\ref{fig:appen_comp_reflection}.
\fsname outperforms ReflectionFlow across diverse categories, including comparison, counting, attributes, and negation, highlighting the advantage of correcting prompts based on shared failure modes rather than relying on per-sample post-hoc edits.

\clearpage

\subsection{Prompt Transferability and Future Work}\label{appen:prompt_generalizability}

\begin{figure*}[ht]
    \centering
    \vspace{-0.2in}
    \includegraphics[width=0.98\linewidth]{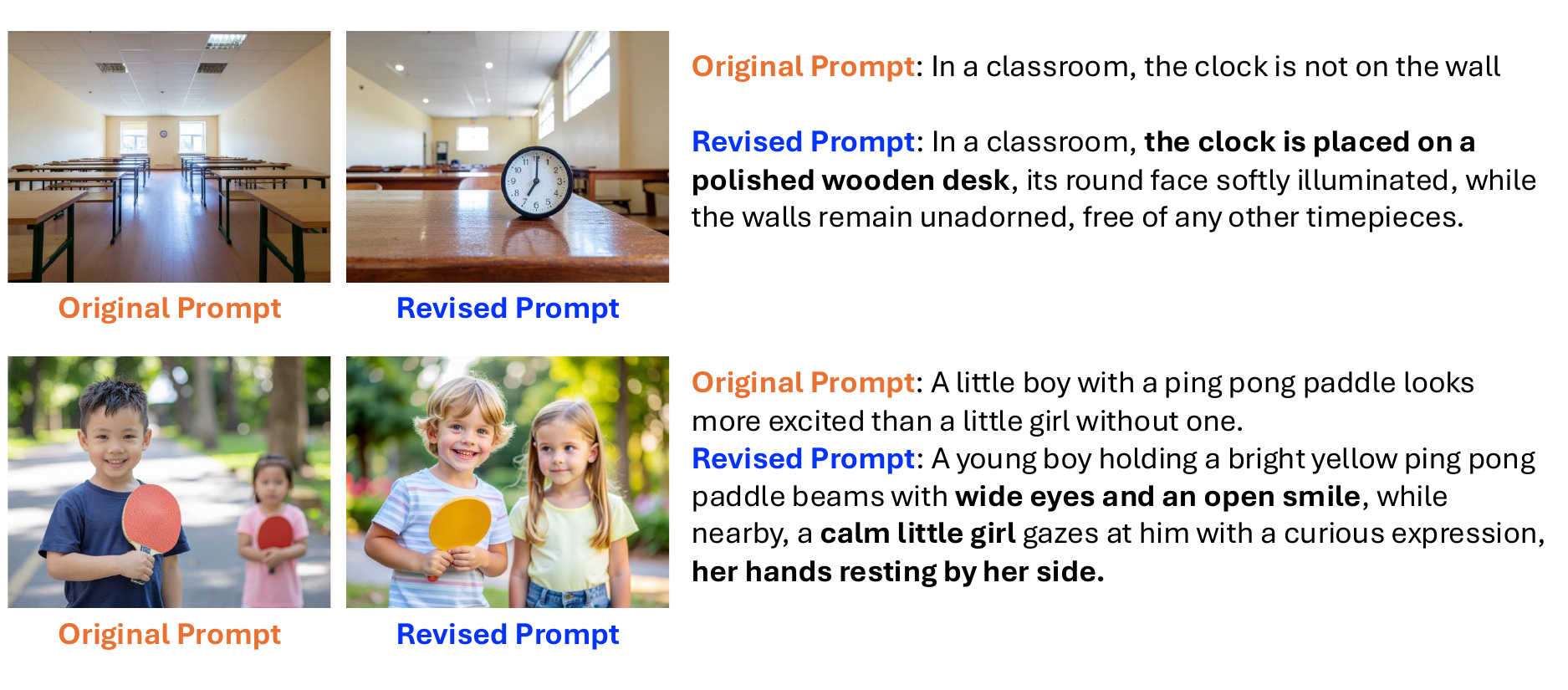}
    \vspace{-0.2in}
    \caption{\textbf{Qualitative example of prompt transferability}. Prompts revised for Flux1.dev are applied to Firefly Image 4 Ultra. By clarifying vague instructions, specifying object presence and absence, and reinforcing contextual cues, the revised prompts yield visuals with stronger adherence compared to those generated from the original prompts.}
    \label{fig:appen_firefly}
    \vspace{-0.05in}
\end{figure*}

We observe that our revised prompts are not only effective for the original generator but also transferable to other models, demonstrating their generalizability. This stems from the fact that our revisions resolve ambiguities in the original prompts, making them more precise and robust. Although different generators may specialize in certain aspects, such as producing fine-grained details or maintaining object counts, they often exhibit overlapping weaknesses. Addressing these weaknesses through prompt revision thus benefits multiple models simultaneously.

Figure~\ref{fig:appen_firefly} illustrates this transferability. The prompts originally revised for Flux1.dev are successfully applied to Firefly Image 4 Ultra. For example, the revised prompts clarify vague or underspecified instructions (e.g., replacing ``not on the wall'' with ``the clock is placed on a polished wooden desk''), making object presence and absence explicit (e.g., reformulating ``the girl is without a ping pong paddle'' into ``her hands resting by her side''), and reinforcing contextual cues. 
These findings suggest a promising research direction: fine-tuning LLMs or other prompt-rewriting systems on pairs of na\"ive user-provided prompts and failure-focused revisions. By learning systematic transformations from short, underspecified, and loosely written prompts into precise, detailed, and effective ones, rather than relying on random expansions, such models could reduce verification costs and inference-time overhead, accelerating the discovery of high-quality prompts from the outset.

\subsection{Future work and limitations.} 
Our core idea---identifying shared failure patterns with high precision and addressing them through targeted prompt revision---is broadly applicable to any text-conditioned generative model. We believe extending this idea to other modalities and tasks is a compelling direction for future research, potentially challenging existing inference-time scaling laws.
In addition, our benchmark provides a new avenue for evaluating verifiers at the attribute level. We also find that prompts refined on one model often generalize well to others (see Appendix~\ref{appen:prompt_generalizability}). This observation suggests a promising direction: fine-tuning LLMs or other prompt-rewriting models using paired data consisting of randomly expanded prompts and their failure-focused revisions. Such training resources could reduce verification overhead and inference-time costs, enabling more efficient discovery of high-quality prompts from the start.

\clearpage
\section{Benchmark Construction and Evaluations}\label{appen:benchmark}

\subsection{Benchmark Category}

\noindent\textbf{Details about benchmark constructions.}
Existing visual evaluation datasets are mostly limited to human-preference annotations. While useful for coarse quality assessment, such datasets are insufficient for our focus: selecting the best-aligned videos from among multiple misaligned candidates, which lies at the core of inference-time scaling.
To address this limitation, we construct a new benchmark explicitly tailored for inference-time scaling and use it to evaluate our verifier, its ablations, and existing baselines. Beyond serving as a testbed for our study, this benchmark also provides a valuable resource for future research on visual prompt-adherence verification.

In our benchmark, each prompt is paired with multiple generated videos, with at least one ground-truth (GT) aligned reference and others containing slight misalignments, thereby forming a mid-quality candidate pool. 
In total, the benchmark comprises 410 prompts. We collect prompts showcased in demos of both popular open-source~\citep{wan2025} and closed-source video models~\citep{veo2025, kling2025}, and categorize them into two broad groups: motion (120 prompts) and physics (144 prompts). To further enrich the evaluation, we also adopt prompts from VBench 2.0, spanning three fine-grained motion-related categories: dynamic attributes (47 prompts), motion order (68 prompts), and motion rationality (31 prompts).
For each prompt, we generate videos using multiple text-to-video models~\citep{kling2025, wan2025, veo2025} as well as image-to-video models~\citep{wan2025}, ensuring the inclusion of both GT-aligned and misaligned outputs. Each video is independently annotated by three human evaluators as GT or non-GT, and the final label is assigned by majority vote.

\noindent\textbf{Detailed analysis of verifiers on our benchmark per category.}
In addition to the overall accuracy reported in Table~\ref{tab:quan_dataset} of the main manuscript, we present per-category accuracy in Table~\ref{tab:appen_verifier_eval}. 
As the results show, \vsname consistently achieves the highest accuracy across all categories. Compared to the decomposed binary VQA baseline, which shares our decomposition strategy but replaces our text-to-text verification with binary VQA, \vsname yields a substantial performance gain, underscoring the advantage of our text-based approach over visual QA methods.
When compared to learned reward models (i.e., MLLM-based verifiers fine-tuned on human-preference datasets), including VideoAlign (the strongest among them and used as our tie-breaker), \vsname still maintains a significant lead. Notably, it achieves this performance without any additional training on preference datasets, but rather through a systematic zero-shot verification process. Furthermore, we attribute this gap to the fact that reward models are typically trained on human-preference data, where subtle aspects such as frame quality, motion smoothness, or stylistic biases often dominate judgments, even when they are not directly related to prompt adherence. In contrast, \vsname focuses explicitly on verifying semantic alignment with the prompt, making it both more accurate and interpretable.

\begin{table*}[h]
\vspace{5pt}
\caption{\textbf{Quantitative results of verifier accuracy per prompt category on our constructed dataset.} \textbf{Bold} indicates the best result.}\label{tab:appen_verifier_eval}
\centering
\vspace{-10pt}
\resizebox{0.98\linewidth}{!}{
\begin{tabular}{l ccccc c}
\toprule

Method   & \Centerstack{Motion} & \Centerstack{Physics} & \Centerstack{Dynamic\\Attributes} & \Centerstack{Motion\\Rationality} & \Centerstack{Motion Order\\Understanding}  & \textbf{\Centerstack{Average}} \\
\midrule
VisionReward~\citep{xu2024visionreward} & 0.650 & 0.569 & 0.319 & 0.662 & 0.452 & 0.571 \\ 
UnifiedReward~\citep{UnifiedReward} & 0.492 & 0.507 & 0.298 & 0.588 & 0.581 & 0.498 \\ 
VideoAlign~\citep{liu2025improving} & \textbf{0.792} & 0.660 & 0.511 & 0.794 & 0.516 & 0.693 \\ 
\midrule
Decomposed binary VQA & 0.733 & 0.667 & 0.617 & 0.809 & 0.613 & 0.700 \\
\fsname (Ours) & \textbf{0.792} & \textbf{0.764} & \textbf{0.638} & \textbf{0.838} & \textbf{0.677} & \textbf{0.763} \\

\bottomrule
\end{tabular}
}
\centering

\end{table*}

While our study focuses on prompt-adherence verification, we believe that our verification framework can be extended to other important axes of evaluation, such as motion quality, NSFW filtering, and bias detection, by replacing prompt decomposition with task-specific decomposition strategies. This flexibility offers promising directions for future research.
\clearpage
\section{Experiments Details}\label{appen:exp_detail}

\subsection{Detailed Setup}
For GenAI-Bench, since many prompts within the same categories (e.g., counting, differentiation, comparison, negation, universal) are similar but differ only in objects, we randomly subsample 20\% to reduce redundancy. 
For selecting $k$, we set $k = N//4$, as $N//2$ samples are first generated for review before prompt revision, and half of them are used as top-performing seeds.

\subsection{Base Model Selection}
To ensure that our study focuses on the effect of prompt redesign in inference-time scaling, we first measure the degree of prompt adherence across candidate leading open-source video models such as Wan, LTX, and Hunyuan. 
This step is necessary because if a model fails to follow the prompt at all, there is little need to apply prompt redesign. 
Specifically, we compute the text embedding similarity between the original prompt and the generated video caption. 
We use Qwen-32B for captioning and employ the SentenceTransformer model (\texttt{intfloat/e5-mistral-7b-instruct}) to measure embedding similarity.
We present the similarity score in Table~\ref{tab:appen_video_model}.

\begin{figure*}[h]
\centering
\captionof{table}{\textbf{Quantitative results of prompt adherence} across different text-to-video models, used to exclude base models with poor alignment and retain only those with acceptable adherence.}
\label{tab:appen_video_model}
\vspace{-0.12in}
\resizebox{0.85\linewidth}{!}{
\begin{tabular}{c l c c c c}
\toprule

Metric & Method   & \Centerstack{Motion\\Rationality} & \Centerstack{Motion Order\\Understanding} & \Centerstack{Dynamic\\Attribute} & \textbf{\Centerstack{Average}} \\
\midrule

\multirow{3}{*}{VideoAlign} & LTX & 0.764 & -0.153 & -0.977 & -0.122 \\
& Hunyuan & 0.904 & 0.212 & -0.775 & 0.114  \\ 
& Wan & \textbf{1.475} & \textbf{0.940} & \textbf{-0.397} & \textbf{0.673}  \\ 
\midrule
\multirow{3}{*}{Text Similarity} & LTX & 0.635 & 0.642 & 0.600 & 0.626 \\
& Hunyuan & 0.678 & 0.671 & 0.616 & 0.655  \\ 
& Wan & \textbf{0.717} & \textbf{0.702} & \textbf{0.631} & \textbf{0.683}  \\

\bottomrule
\end{tabular}
}
\end{figure*}

Based on this analysis, we selected Wan as our primary video model, since it demonstrates a reasonable level of prompt adherence while leaving room for improvement through verification and redesign. 
In contrast, models such as LTX and Hunyuan were excluded, as their low adherence made them unsuitable for evaluating prompt redesign at inference-time scaling, particularly on complex prompts in VBench2.0 that involve status changes or multiple consecutive events within a single video.

\clearpage
\section{Additional Qualitative Experimental Results}\label{appen:more_results}

\subsection{Text-to-Image Generation}
\begin{figure}[h]
    \centering
    \vspace{-0.15in}
    \includegraphics[width=0.63\linewidth]{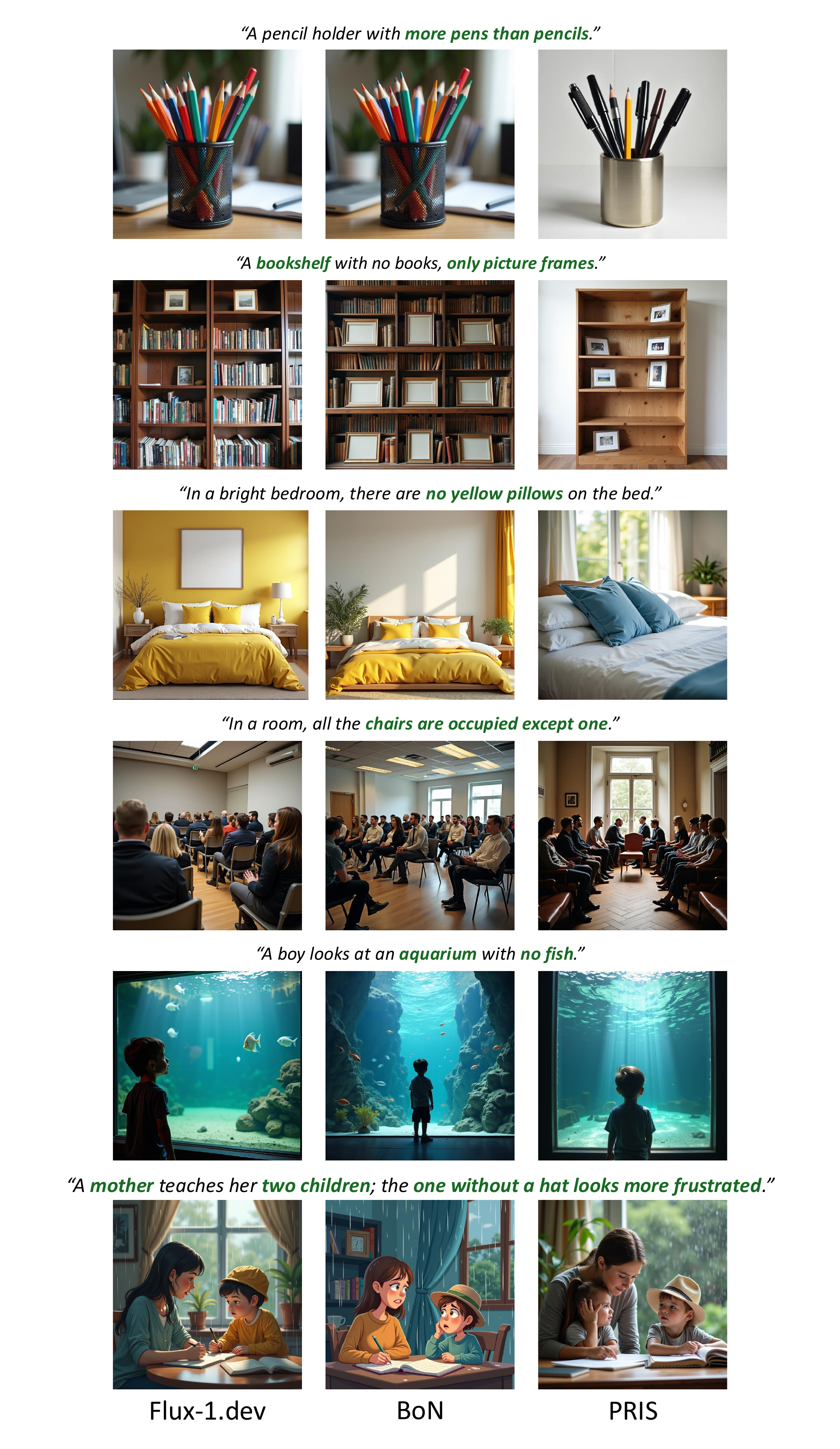}
    \vspace{-0.2in}
    \caption{\textbf{Qualitative comparisons on T2I generation} where visual generation is (initially) conditioned on the original prompts.}
    \label{fig:appen_bon}
    \vspace{-0.2in}
\end{figure}

We provide additional qualitative results beyond Figure~\ref{fig:qual_flux}, demonstrating that our prompt redesign improves coherence of the final visual outputs under the same NFE budget (2000, as in the main experiments). 
As shown in Figure~\ref{fig:appen_bon}, which compares the top-scoring outputs generated from the original GenAI-Bench prompts, our method performs particularly well on prompt sets containing ambiguous attributes, numerical specifications, or subtle constraints (e.g., ``without,'' ``greater variety''), effectively elaborating them into more faithful visual realizations than baselines.

\begin{figure}[h]
    \centering
    \vspace{-0.2in}
    \includegraphics[width=0.65\linewidth]{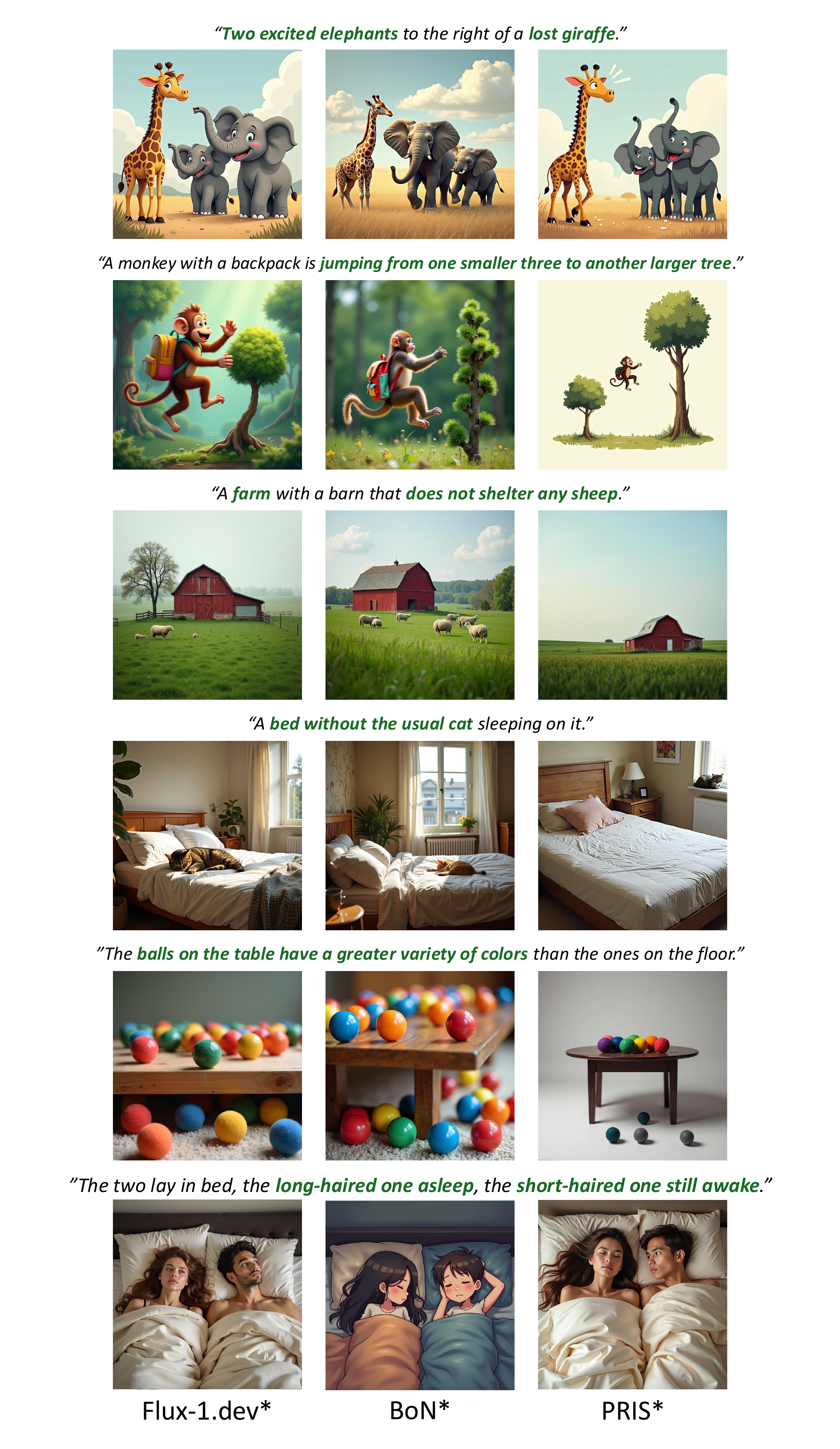}
    \vspace{-0.2in}
    \caption{\textbf{Qualitative comparisons on T2I geneation} where visual generation is (initially) conditioned on standard prompt expansion.}
    \label{fig:appen_bon*}
    \vspace{-0.2in}
\end{figure}

We also compare with standard prompt expansion in Figure~\ref{fig:appen_bon*}, where ours achieves substantially higher prompt fidelity compared to the baselines.
Unlike standard prompt expansion, which cannot target or identify the most challenging semantic elements, our joint scaling of visuals and prompts more faithfully preserves the intended semantics by adaptively revising the prompt based on recurring failure modes.

\clearpage

\subsection{Text-to-Video Generation}
\begin{figure}[ht]
\vspace{-0.1in}
\includegraphics[width=0.95\textwidth]{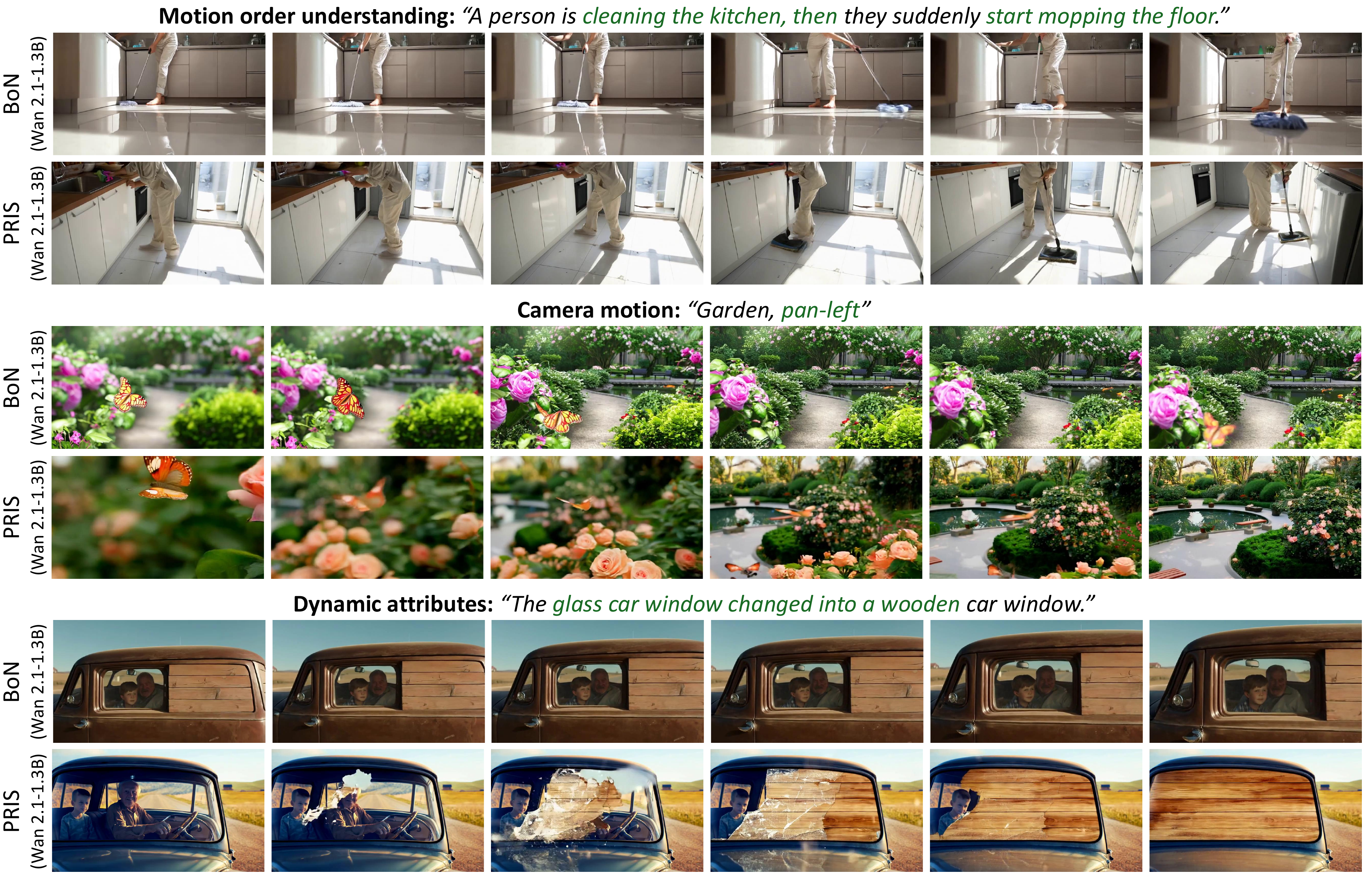}
\vspace{0.5in}
\includegraphics[width=0.95\textwidth]{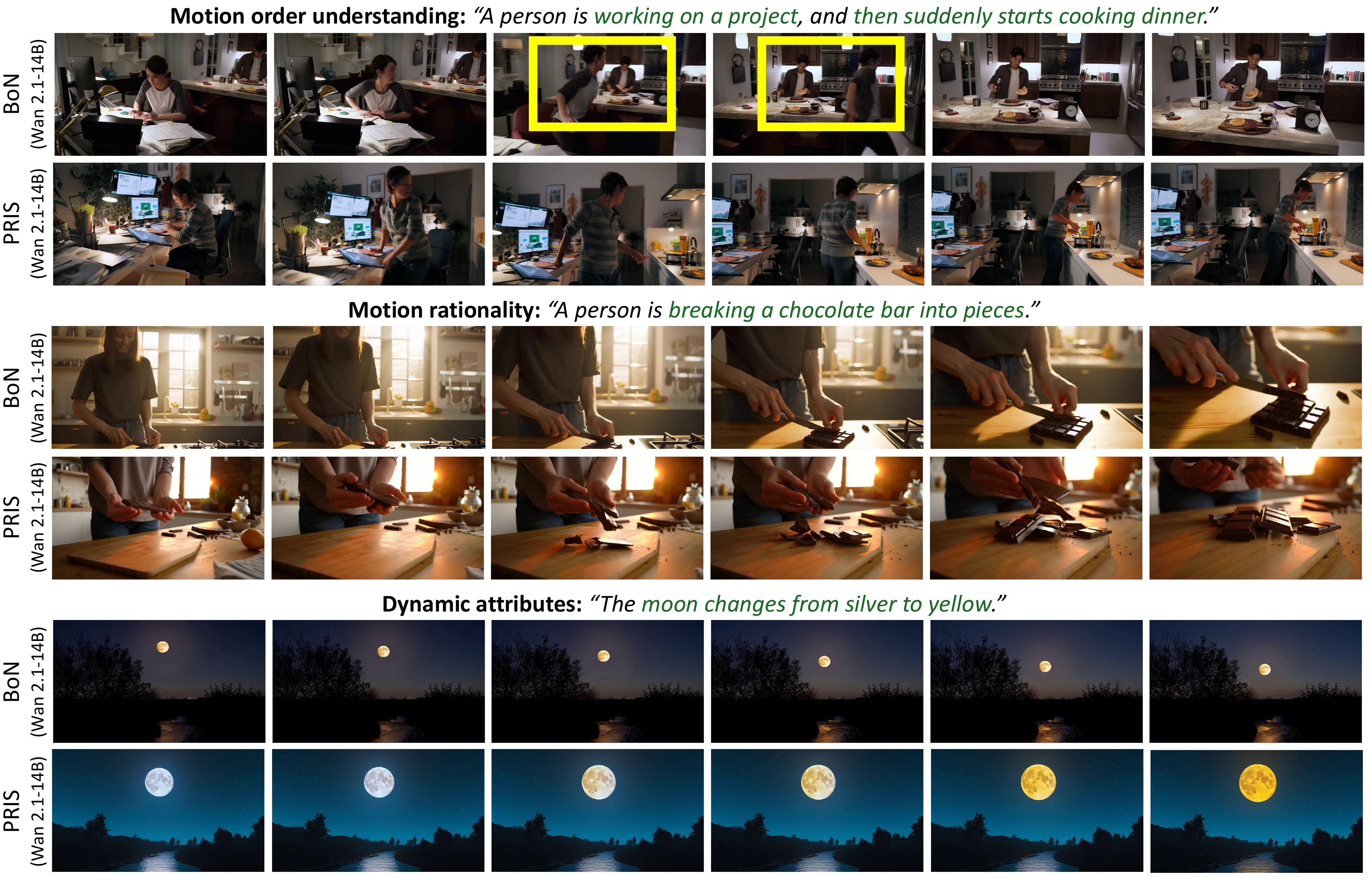}
\vspace{-0.5in}
\captionof{figure}{\textbf{Qualitative comparisons on T2V generation} where visual generation is (initially) conditioned on standard prompt expansion, with Wan2.1-1.3B (top) and Wan2.1-14B (bottom).
}\label{fig:appen_vbench2.0}
\vspace{-0.2in}
\end{figure}

In addition to Figure~\ref{fig:vbench2.0}, we present additional qualitative top-scoring examples in Figure~\ref{fig:appen_vbench2.0}. As shown, our method more faithfully follows the intent of the original prompt. The final top-scoring visuals generated with our \fsname demonstrate significantly stronger prompt adherence compared to baselines. Specifically, BoN often misses key events or produces unnatural temporal order. For example, it may depict only a single motion (e.g., morphing without differentiating ``cleaning the kitchen'' in the 1st visual) or assign different motions to different people (in the 4th visual).
BoN also frequently fails to capture dynamic changes, generating only static states (3rd and 6th visuals).
Furthermore, BoN often does not correctly realize sequential actions, such as repeatedly attempting to break chocolate pieces, whereas our method generates coherent sequences where the person both attempts the action and displays the broken pieces (5th visual).

\subsection{More Visualizations}\label{appen:more_vis}
We include an HTML file to the attached zip file. To explore the generated visuals and comparisons with baselines alongside their corresponding prompts, please open \texttt{visuals/index.html} in a Chrome browser (This file is located in the \texttt{visuals} directory within the attached zip file). This visualizes the generated visuals, including images and videos, in the \texttt{visuals/resources} folder.

\end{document}